\documentclass[letterpaper]{article} 
\usepackage{aaai24}  
\usepackage{times}  
\usepackage{helvet}  
\usepackage{courier}  
\usepackage[hyphens]{url}  
\usepackage{graphicx} 
\usepackage{natbib}  
\usepackage{caption} 
\usepackage{algorithm}
\usepackage{algorithmic}

\usepackage{newfloat}
\usepackage{listings}

\usepackage{epsfig}
\usepackage{amsmath}
\usepackage{multirow}
\usepackage{bbding}
\usepackage{amsfonts}
\usepackage{booktabs}

\DeclareCaptionStyle{ruled}{labelfont=normalfont,labelsep=colon,strut=off} 
\floatstyle{ruled}
\newfloat{listing}{tb}{lst}{}
\floatname{listing}{Listing}
\title{Learning Explicit Contact for Implicit Reconstruction of Hand-Held Objects from Monocular Images}

\author{
    Junxing Hu\textsuperscript{\rm 1,2},
    Hongwen Zhang\textsuperscript{\rm 3},
    Zerui Chen\textsuperscript{\rm 4},
    Mengcheng Li\textsuperscript{\rm 5},\\
    Yunlong Wang\textsuperscript{\rm 2},
    Yebin Liu\textsuperscript{\rm 5},
    Zhenan Sun\textsuperscript{\rm 1,2}\thanks{Corresponding author.}
}

\affiliations{
    \textsuperscript{\rm 1}School of Artificial Intelligence, University of Chinese Academy of Sciences\\
    \textsuperscript{\rm 2}State Key Laboratory of Multimodal Artificial Intelligence Systems, Institute of Automation, Chinese Academy of Sciences\\
    \textsuperscript{\rm 3}School of Artificial Intelligence, Beijing Normal University\\
    \textsuperscript{\rm 4}Inria, DI ENS, CNRS, PSL Research University\\
    \textsuperscript{\rm 5}Tsinghua University\\
    junxing.hu@cripac.ia.ac.cn, znsun@nlpr.ia.ac.cn
}

\begin{document}

\maketitle


\begin{abstract}
Reconstructing hand-held objects from monocular RGB images is an appealing yet challenging task. In this task, contacts between hands and objects provide important cues for recovering the 3D geometry of the hand-held objects. Though recent works have employed implicit functions to achieve impressive progress, they ignore formulating contacts in their frameworks, which results in producing less realistic object meshes. In this work, we explore how to model contacts in an explicit way to benefit the implicit reconstruction of hand-held objects. Our method consists of two components: \emph{explicit contact prediction} and \emph{implicit shape reconstruction}. 
In the first part, we propose a new subtask of directly estimating 3D hand-object contacts from a single image.
The part-level and vertex-level graph-based transformers are cascaded and jointly learned in a coarse-to-fine manner for more accurate contact probabilities. 
In the second part, we introduce a novel method to diffuse estimated contact states from the hand mesh surface to nearby 3D space and leverage diffused contact probabilities to construct the implicit neural representation for the manipulated object. Benefiting from estimating the interaction patterns between the hand and the object, our method can reconstruct more realistic object meshes, especially for object parts that are in contact with hands. Extensive experiments on challenging benchmarks show that the proposed method outperforms the current state of the arts by a great margin.
Our code is publicly available at \textit{https://junxinghu.github.io/projects/hoi.html}.
\end{abstract}

\section{Introduction}
Reconstructing human-object interaction from monocular images is essential to understand the interactions between humans and the physical world. 
Toward this goal, recent progress has been achieved in the individual reconstruction of the body~\cite{kocabas2021pare,zhang2020learning,zhang2023pymaf}, hands~\cite{MANO:SIGGRAPHASIA:2017,Kulon2020weaklysupervisedmh,baek2019pushing,hampali2022keypoint,chen2021camera,boukhayma20193d,li2022interacting}, objects~\cite{chen2019learning,mescheder2019occupancy,park2019deepsdf,groueix2018papier,wang2018pixel2mesh,peng2021shape}, and their joint reconstruction~\cite{hasson2019learning,hasson2020leveraging,karunratanakul2020grasping,yang2021cpf,chen2022alignsdf,chen2023gsdf,ye2022s}. 
However, this task remains very challenging due to the complexity of hand poses and the diversity of interacting objects.

As hand-held objects involve the grasp configuration between hands and objects, the contacts play essential roles in modeling their interactions.
To improve the interaction, current methods model the contact in different representations, including using contacts to optimize meshes~\cite{hasson2019learning}, the contact potential field~\cite{yang2021cpf}, or the grasping field~\cite{karunratanakul2020grasping}. 
However, these methods only model contacts as an additional loss function, which miss the chance to construct and exploit contact priors to simplify the 3D reconstruction problem.

\begin{figure}[t]
\centering
\includegraphics[width=1.0\columnwidth]{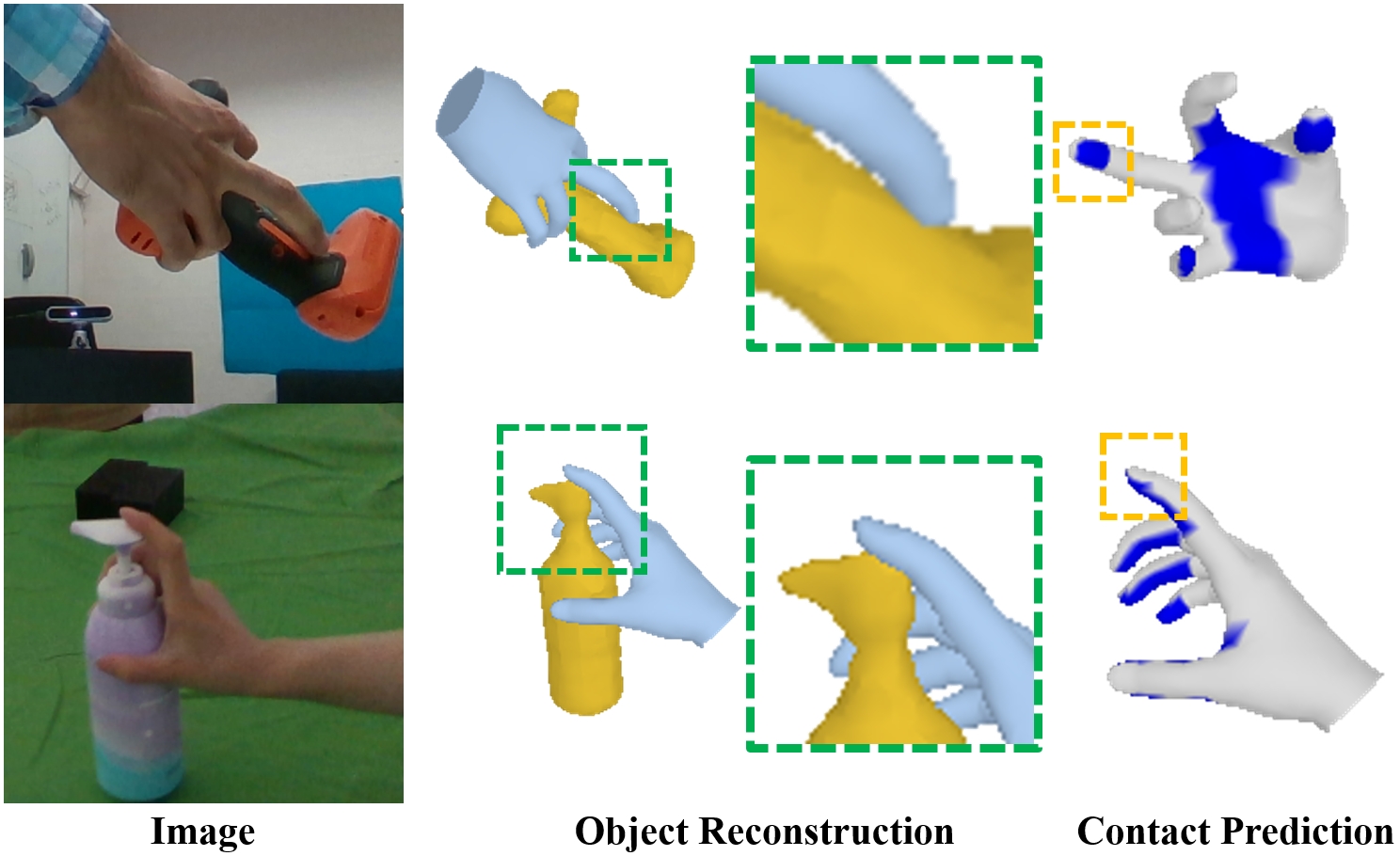}
\caption{Given an RGB image, the proposed method predicts hand-object contacts and recovers the 3D geometry of the object. The insight is that the contacts could provide effective cues for the hand-held object reconstruction.}
\label{first_fig}
\end{figure}

Our key observation is that the contacts between hands and objects provide important cues for recovering the hand-held object. Modeling contacts between them can compensate for the lack of 3D information in monocular images and makes it easier to infer the shape of the object, especially for parts that are in contact with hands as shown in Fig.~\ref{first_fig}. 
Though previous methods have included contact losses~\cite{hasson2019learning,karunratanakul2020grasping} or optimization objectives~\cite{yang2021cpf,grady2021contactopt}, they do not consider the usage of contacts as an intermediate representation to benefit the 3D reconstruction.
In this work, we explore how to construct contact priors from the monocular RGB image to help recover the 3D object geometry. Specifically, we first predict contact points explicitly on the hand mesh surface.
To our knowledge, estimating contact states from a single RGB image is explored only for human body mesh~\cite{huang2022capturing,fieraru2020three,fieraru2021learning,chen2023detecting} without focusing on the hand.
To this end, we introduce a novel coarse-to-fine learning framework to jointly learn part- and vertex-level contact states. In addition, we utilize the graph-based transformer which combines graph convolutions with transformers to accumulate relevant features among adjacent nodes in the hand mesh and obtain robust contact predictions.

Then, we attempt to exploit predicted contact states to simplify the 3D reconstruction task. Here, we follow the previous work~\cite{ye2022s} to model hand-held objects with deep implicit functions~\cite{park2019deepsdf}, which can generate realistic and high-resolution object meshes. However, how to make implicit functions take good advantage of estimated contact states is also challenging and remains unsolved. The main challenge is that contact points are distributed on the hand surface in the discrete form, while implicit functions have continuous values in the whole 3D volume. To tackle the difficulty, we employ sparse convolutions to diffuse these discrete contact states from the hand surface to the 3D space. Then, the implicit function can naturally query corresponding contact features for a given 3D point and improve the neural implicit reconstruction. We conduct extensive experiments on HO3D~\cite{hampali2020honnotate} and OakInk~\cite{yang2022oakink} benchmarks to show that our method can reconstruct high-quality object meshes that interact faithfully with hands.

To sum up, the main contributions can be listed as follows:
\begin{itemize}
\item We propose to leverage contact priors for better reconstruction of hand-held objects. To estimate contact states more accurately, we introduce a novel framework that jointly improves part-level and vertex-level contact states in a coarse-to-fine manner.
\item To make discrete contact states compatible with continuous implicit shape functions, we propose to diffuse contact features from the hand mesh surface to the whole 3D volume, which enables the continuous query of contact features for implicit object reconstruction.
\item We conduct extensive experiments on HO3D and OakInk benchmarks to validate the effectiveness of our method. Our method can produce more realistic hand-held object meshes and advance state-of-the-art accuracy.
\end{itemize}


\section{Related Work}
Our work focuses on reconstructing hand-held objects from monocular RGB images. In this section, we first review related works in the field of 3D hand-object reconstruction. Then, we discuss how to leverage contact information to improve the quality of 3D reconstruction. 

\textbf{3D Hand-object Reconstruction}. This task aims to reconstruct the 3D geometry of hands and hand-held objects from images. Existing approaches can be generally classified into two categories: multi-view and single-view methods. Multi-view methods~\cite{hampali2020honnotate,yang2022oakink,chao2021dexycb,oikonomidis2011full,wang2013video} employ multiple cameras positioned at different viewpoints to infer the 3D structure of the grasping scenario. Though this type of method can generate very accurate 3D reconstruction results, they need careful camera calibrations and are inconvenient to deploy in the wild scene. Single-view methods only need monocular sensors~\cite{ye2022s,hasson2019learning,hasson2020leveraging,yang2021cpf,karunratanakul2020grasping,chen2022alignsdf,chen2023gsdf,tse2022collaborative,zhang2021single,hu2022physical,kyriazis2014scalable,zhao2022stability} as inputs and are flexible to apply in real practice. In this work, we use the most common monocular RGB images as inputs. However, given the ill-posed nature, it is quite challenging to infer the 3D structure only from monocular RGB cues. To alleviate the difficulty of the hand reconstruction problem, Hasson \emph{et al.}~\cite{hasson2019learning,hasson2020leveraging} propose to employ the parametric hand model MANO~\cite{MANO:SIGGRAPHASIA:2017}, which encodes rich hand priors, to predict the hand mesh. To produce more realistic hand meshes, recent works~\cite{karunratanakul2020grasping,chen2022alignsdf,chen2023gsdf} employ the neural implicit function~\cite{park2019deepsdf} to model the hand shape and use estimated hand pose priors~\cite{chen2022alignsdf,chen2023gsdf} to simplify the hand shape learning. However, compared with the hand part, hand-held object reconstruction is even more challenging. Since there are thousands of manipulated objects in our daily lives, it is difficult to make a unified object mesh template like MANO or estimate 6D poses reliably for diverse objects, especially for symmetric objects. Given its difficulty, some existing works~\cite{yang2021cpf,hasson2020leveraging,yang2022artiboost} even make a strong assumption that the perfect object model is known at test time and only predicts its 6D pose. A recent work~\cite{ye2022s} relaxes this assumption and proposes to leverage estimated hand poses to benefit the model-free reconstruction of hand-held objects. In this work, we go a step further and argue that contacts between hands and objects could provide important cues for 3D reconstruction and introduce a novel framework to generate more realistic object meshes that interact with hands.

\begin{figure*}[t]
\centering
\includegraphics[width=1.0\textwidth]{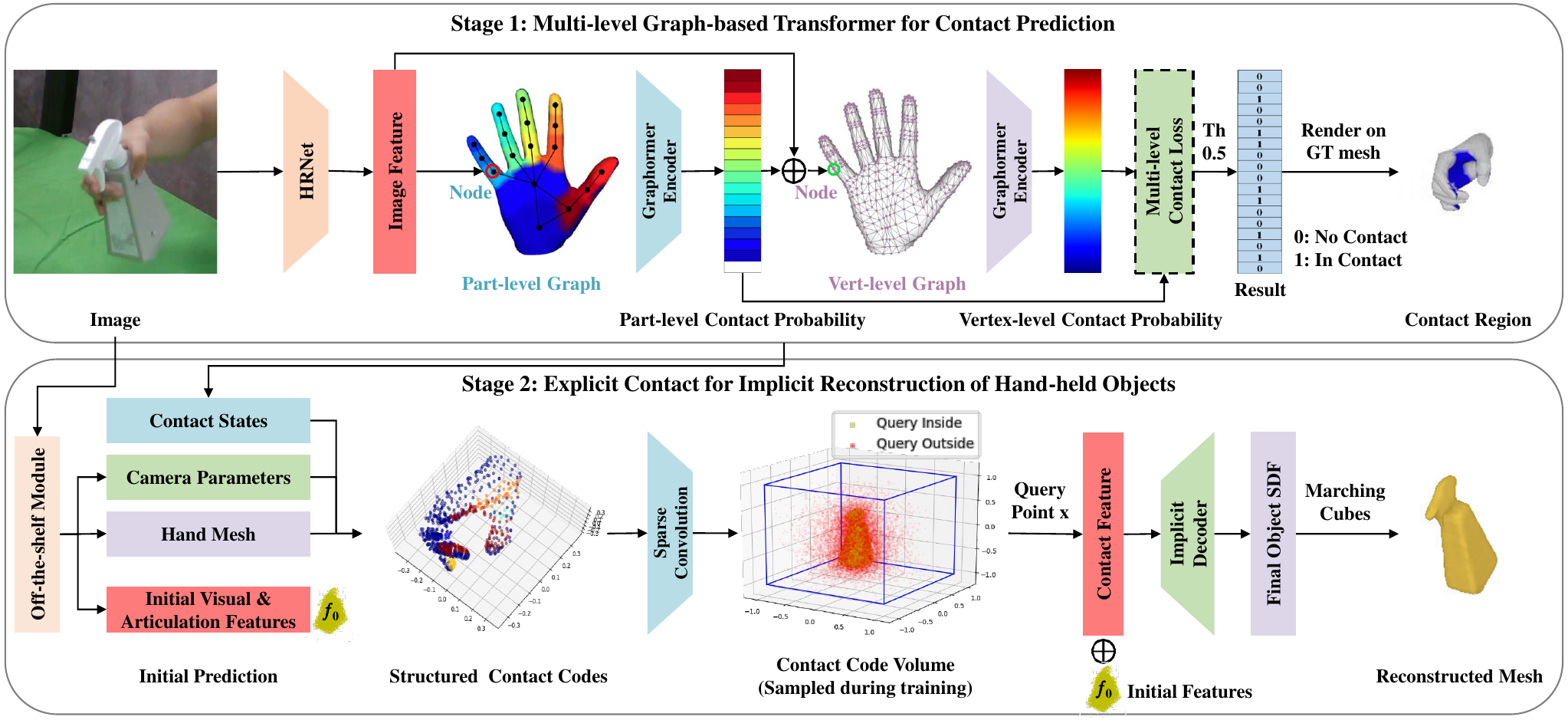} 
\caption{The overview of learning explicit contact for implicit reconstruction.
First, the method estimates hand contact regions given a monocular RGB image. Based on the template hand mesh, part- and vertex-level graph-based transformers are cascaded for accurate predictions.
Second, the estimated contact is used to construct the implicit neural representation. An off-the-shelf module is utilized to produce the camera parameters, hand mesh, and initial features.
Then, the structured contact codes are generated by anchoring contact probabilities to the hand mesh surface.
After sparse convolutions, the contact states on the hand surface are diffused to its nearby 3D space, which facilitates the perception and reconstruction of the manipulated object.}
\label{pipeline}
\end{figure*}

\textbf{Contacts in Object Reconstruction and Manipulation}. 
Learning to model and reconstruct the 3D geometry of objects from monocular images has been a crown jewel in the field of computer vision~\cite{roberts_thesis_1963,mundy2006geometric}. Previous works usually represent the 3D object using explicit representations (\emph{e.g.}, meshes~\cite{groueix2018papier,wang2018pixel2mesh}, point clouds~\cite{qi2017pointnet,qi2017pointnet++} or voxels~\cite{choy20163d,Riegler2017OctNet,pavlakos2017coarse}) and use deep neural networks to predict them. In recent years, neural implicit functions~\cite{park2019deepsdf,mescheder2019occupancy,chen2019learning} have gradually become a popular paradigm for 3D reconstruction. It is seamlessly compatible with neural networks and can theoretically reconstruct objects at unlimited resolution. However, the implicit function itself does not contain object surface priors, which makes it hard to fit diverse object surfaces. In this work, we construct a surface prior using contacts between hands and objects and simplify the learning problem. Actually, some works in robotics~\cite{bicchi2000robotic,li2020review,buescher2015augmenting,touch-dexterity,tse2022s} have shown that contacts could provide rich cues about the object shape and how to manipulate the given object. Some recent systems~\cite{touch-dexterity,jain2019learning} can successfully manipulate different objects by using contact sensors. However, how to use contacts to benefit hand-held reconstruction is under-explored in our task. Previous works only use contact information in an implicit way. Methods using explicit object representations~\cite{hasson2019learning,yang2021cpf,grady2021contactopt} introduce contact loss terms to encourage objects to be close to reconstructed hand meshes. Some recent efforts~\cite{karunratanakul2020grasping,ye2022s} also introduce contact loss terms in the context of neural implicit representation. 
Different from them, we model and predict contact states explicitly and successfully leverage volume encoding~\cite{peng2021neural,kwon2021neural,choi2022mononhr} to diffuse contact information from hand surfaces to the 3D space for the hand-held object reconstruction. 


\section{Method}
In this section, we describe the technical details of the proposed method.
As shown in Fig.~\ref{pipeline}, our method consists of two stages: explicit contact prediction and implicit shape reconstruction. In the first stage, we propose to predict part-level and vertex-level hand contact states in a coarse-to-fine manner. A graph-based transformer model is introduced to estimate contact probabilities more accurately. In the second stage, we present a novel method to leverage estimated contact states to improve the neural implicit reconstruction of hand-held objects.

\subsection{Explicit Contact Prediction}\label{stage1}
Given a single RGB image $I$, our method first predicts the contact regions between the hands and objects. Specifically, we estimate contact probabilities within $[0, 1]$ on hand meshes to measure the likelihood of the region touching the object.
In our method, the contact probabilities are predicted from coarse to fine and denoted as $\boldsymbol{C_{p}}=\{\boldsymbol{c_{p}}^i \in [0, 1]\}_{i=1}^{N_{p}}$ and $\boldsymbol{C_{v}}=\{\boldsymbol{c_{v}}^i \in [0, 1]\}_{i=1}^{N_{v}}$ for the part-level and vertex-level contacts, where $N_{p}$ and $N_{v}$ are the number of the hand parts and hand mesh vertices, respectively.

\paragraph{Multi-level Contact Graphs.}
For more accurate predictions of the contact probabilities, multi-level contact graphs are leveraged to process the surface regions in the part and vertex levels such that the contact can be jointly learned from coarse to fine.
Considering that the hand mesh can be naturally represented as a graph, we build the contact graphs based on the template MANO mesh~\cite{MANO:SIGGRAPHASIA:2017}.
Specifically, the part-level graph $G_{p}$ with $N_{p}$ nodes is generated relying on a coarse division of the hand regions.
According to statistical contact frequency, the hand surface is divided into $N_{p}$ subregions, including ($N_{p}-1$) subregions on the hand palm and one subregion on the back side of the hand.
When building graph $G_{p}$, the center point of each part of the MANO template is taken as a graph node.
For each graph node, its features are the concatenation of the image-based feature and its 3D coordinates.
As shown in the first stage in Fig.~\ref{pipeline}, an image feature $f \in \mathbb{R}^{D}$ with the length of $D$ is extracted from $I$ by using an HRNet backbone~\cite{wang2020deep}.
Therefore, each part-level graph node feature of $G_{p}$ is $\boldsymbol{g_{p}^i} \in \mathbb{R}^{D+3}, i=\{1,2,\dots,N_{p} \}$ and the adjacency matrix is encoded as the physical contact relationship between nodes.
On the other hand, the vertex-level graph $G_{v}$ is generated based on the $N_{v}$ mesh vertices with an adjacency matrix from the MANO template.
In addition to the image feature, the vertex-level node features of $G_{v}$ also include the part-level contact probability $\boldsymbol{C_{p}}$, resulting in the node feature $\boldsymbol{g_{v}^i} \in \mathbb{R}^{D+N_{p}+3}, i=\{1,2,\dots,N_{v}\}$.

\paragraph{Graph-based Transformer for Contact Prediction.}
In hand-object interaction, the contact area is usually occluded by hands or objects, which requires the network to perceive local details and global information.
Following Graphormer~\cite{lin2021mesh}, our contact estimators are designed as graph-based transformers that incorporate the graph convolution~\cite{kipf2017semi} into the transformer block~\cite{Vaswani2017Attention}.
In this way, the graph convolution focuses on fine-grained local interactions, while the latter encodes the global relationships of the whole hand regions.
As the contacts are predicted at the part and vertex levels, the architectures of the coarse and fine contact estimators are also built upon the graphs $G_{p}$ and $G_{v}$, respectively.
Specifically, the coarse and fine contact estimators have $N_p$ and $N_v$ input tokens, which correspond to the same number of nodes in the graphs.
Moreover, the two contact estimators have different hidden sizes in their transformer blocks.
In practice, we find that a hidden size of 256 is sufficient for the part-level contact estimation, and the three blocks with hidden sizes of 1024, 256, and 64 work well for the vertex-level contact estimator.

For both the two contact estimators, the size of the output token is set to one. 
Similar to the settings in BSTRO~\cite{huang2022capturing}, a sigmoid function is used to convert output tokens to contact probabilities in the range of [0, 1], and we extract contact points with probabilities greater than 0.5.

\subsection{Explicit Contact for Implicit Object Reconstruction}\label{stage2}
As shown in the second stage in Fig.~\ref{pipeline}, given the explicit contact prediction $\boldsymbol{C_{p}}$ and $\boldsymbol{C_{v}}$ with the hand mesh, our method first builds structured contact codes in a normalized 3D space. 
Then, they are fed into a sparse convolutional network to generate the contact code volumes $\boldsymbol{V}$ at different resolutions.
This operation diffuses the contact states on the hand surface to the nearby 3D space and can be sampled continuously as additional conditions for the implicit reconstruction of objects.

\paragraph{Initial Prediction.}
Given an RGB image, an off-the-shelf module from IHOI~\cite{ye2022s} is used to generate the camera parameters, the hand mesh, and initial features $f_{0}$ including visual and articulation embeddings.
By using the camera parameters, sampled 3D query points on the object surface are transformed into a normalized coordinate system around the hand wrist, which serve as the inputs for the subsequent structured contact codes.

\paragraph{Structured Contact Codes.}
The predicted contact states $\boldsymbol{C_{v}} \in \mathbb{R}^{N_{v}}$ are utilized to construct structured contact codes, which act as intermediate contact features. In the context of implicit reconstruction, we perform trilinear interpolation on estimated contact probabilities according to the contact point's position.
In addition, to facilitate the network learning, each contact code $\boldsymbol{c_{v}^i} \in \mathbb{R}^{1}$ is mapped to a higher dimensional space by using the positional encoding ~\cite{mildenhall2021nerf}.

\paragraph{Contact Code Volume.}
There are two disadvantages of directly extracting features from structured contact codes.
First, the contact information is only limited to the mesh surface and cannot cover the surrounding space of the hand where the object is located.
Second, the vertices are too sparse in 3D space to provide enough contact information as most extracted features are zero vectors.
Since the implicit functions have continuous values in the 3D volume, the sparse convolutions~\cite{graham20183d} are utilized to diffuse the discrete contact states to the continuous space. 
Specifically, the structured contact codes are first scaled into the initial volume $V_{0}$ as the input.
Then, a sparse convolutional network is used to process the contact code volumes $\boldsymbol{V} = \{V_{i}\}_{i=1}^{L}$ at $L$ different resolutions inspired by Neural Body~\cite{peng2021neural}.
As a result, the contact code volumes are not limited to contact states at the hand mesh surface and contain diffused contact features for nearby 3D space, which is compatible with the continuous implicit functions.

\paragraph{Implicit Decoding.}
Contact code volumes of different resolutions are first normalized to the same scale [-1, 1]. Then, the contact feature $fc_{i}$ is extracted by interpolation according to the query point $x$ from each contact code volume $V_{i}$. The final contact feature $fc$ is obtained as the concatenation of features extracted from volumes of different resolutions: 
\begin{equation}
  fc = \bigoplus(fc_{1}, fc_{2}, \cdots, fc_{L})
\end{equation}
where $ \bigoplus(\cdot) $ is a concatenation operation. 
After that, the SDF value $s$ on the query point $x$ can be computed via an implicit function $\mathcal{F}$ given the conditions of the contact feature $fc$ and the initial features $f_{0}$: 
\begin{equation}
  s = \mathcal{F}(x, fc, f_{0}) \label{sdf_formula}
\end{equation}
Similar to other methods~\cite{ye2022s, chen2022alignsdf}, the implicit function $\mathcal{F}$ is implemented as a decoder network similar to DeepSDF~\cite{park2019deepsdf}, which composes of eight fully connected layers with a skip connection at the fourth layer.

\subsection{Training Details}\label{trainingDetails}

\begin{figure}[t]
\centering
\includegraphics[width=0.8\columnwidth]{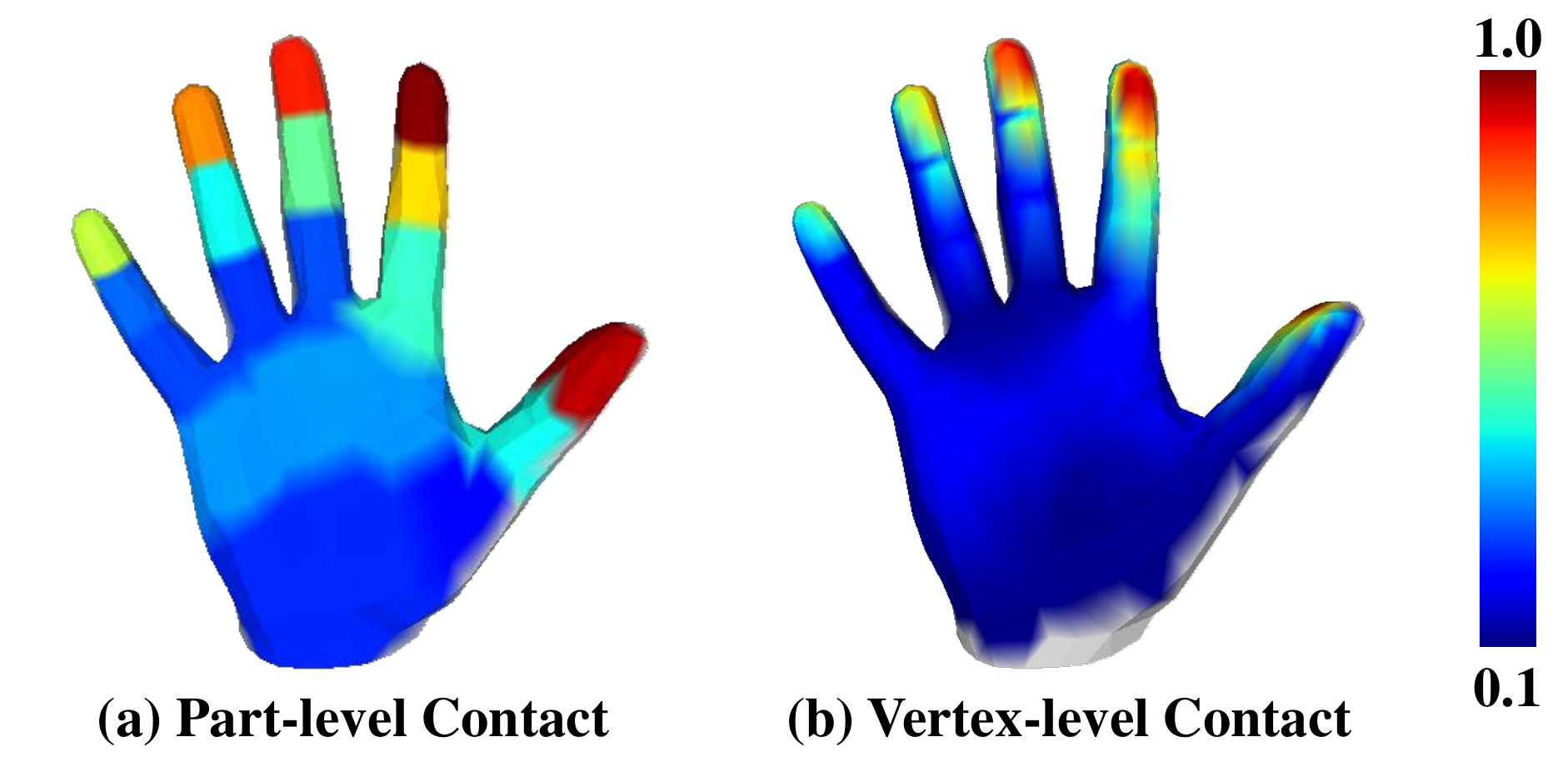}
\caption{Visualization of contact frequency for different hand regions on OakInk~\cite{yang2022oakink}. (a) Part-level contact. (b) Vertex-level contact.}
\label{contact_frequency}
\end{figure}

\paragraph{Contact Prediction.}
In the first stage, the framework is trained in an end-to-end fashion to estimate the contact region from a single image. 
During training, the loss $\mathcal{L}_{Contact}$ is used as follows:
\begin{equation}
\mathcal{L}_{Contact} = \lambda_{p} \mathcal{L}_{part} + \lambda_{v} \mathcal{L}_{vertex} + \lambda_{vs} \mathcal{L}_{vertex\_sub}
\end{equation}
where $\lambda_{p}$, $\lambda_{v}$, and $\lambda_{vs}$ are balancing weights.
$\mathcal{L}_{part}$, $\mathcal{L}_{vertex}$, and $\mathcal{L}_{vertex\_sub}$ are weighted binary cross entropy (BCE) losses between the ground truth and the predicted contact probabilities.
The first one corresponds to the part-level contact.
For multi-scale perception and computational efficiency, the template MANO mesh is downsampled to a sub-mesh with 195 vertices for graph generation.
Processed by graphormer encoders, the coarse contact prediction is generated to compute the $\mathcal{L}_{vertex\_sub}$. 
Then, the coarse prediction is upsampled back to 778-dimensional refined results for $\mathcal{L}_{vertex}$.
Fig.~\ref{contact_frequency} illustrates the contact frequency on different hand regions by analyzing statistics on a large hand-object interaction dataset OakInk~\cite{yang2022oakink}.
It can be observed that the frequencies of different regions vary greatly. Therefore, we normalize these frequencies to $[0.1, 1]$ and use them as weight priors to compute the weighted BCE losses.

\paragraph{Implicit Reconstruction.}
In the second stage, the off-the-shelf module from IHOI~\cite{ye2022s} is trained together with the proposed method.
Similar to IHOI, the object model is provided to guide query point sampling during training, where 95\% of the points are sampled around the model surface and others uniformly in the normalized space as shown in Fig.~\ref{pipeline}. 
It should be noted that at test time, query points are uniformly sampled in space since the object model is agnostic.
In this part, the reconstruction loss $\mathcal{L}_{Recon}$ is calculated as follows:
\begin{equation}
\mathcal{L}_{Recon} = \mathcal{L}_{obj} + \mathcal{L}_{hoi}= \left\|s-\hat{s}\right\|_1 + \frac{1}{N_c} \sum_{i=1}^{N_c}({c_{v}^i} \cdot |s_{h}^i|)
\end{equation}
where $\mathcal{L}_{obj}$ is an L1 loss function between the ground truth $\hat{s}$ and the predicted SDF value $s$ of the object similar to other approaches~\cite{ye2022s, chen2022alignsdf, chen2023gsdf}.
$\mathcal{L}_{hoi}$ is related to ${N_c}$ vertices on the hand mesh that are in contact with the object (\emph{i.e.}, the contact probability $c_{v}^i >0.5$).
$s_{h}^i$ is the SDF value calculated in Equation~\ref{sdf_formula} of the hand contact vertices.
Taking $c_{v}^i$ as the weight, $\mathcal{L}_{hoi}$ is the weighted average sum of the SDF values. This term serves as a regularization term to penalize hand contact points that penetrate or are far from the object.

\section{Experiments}\label{EXPERIMENT}

\subsection{Implementation Details}
In this work, the size of the hand-object centered image is $ 224 \times 224$.
The number of graph nodes are $N_{p}=18$ and $N_{v}=778$. 
The length of the image feature is $D=2048$. 
The shapes of contact code volumes ($L=4$) are $V_{0}=[64,~64,~64],~V_{1}=[32,~32,~32],~V_{2}=[16,~16,~16],~V_{3}=[8,~8,~8],~V_{4}=[4,~4,~4]$, and their code dimensions are $d_{0}=16,~d_{1}=32,~d_{2}=64,~d_{3}=d_{4}=128$.
The balancing weights are $\lambda_{p}=1,~\lambda_{v}=\lambda_{vs}=0.5$. 
The model is implemented by PyTorch~\cite{paszke2019pytorch} and the HRNet backbone~\cite{wang2020deep} is pre-trained on ImageNet~\cite{wang2020deep}.
The learning rate is set to 1e-4, and the Adam optimizer~\cite{KingmaB14} is used.
Each model is trained for 200 epochs on the RTX3090 GPU.

\subsection{Datasets and Setup}
The proposed method is evaluated on two challenging real-world datasets: OakInk~\cite{yang2022oakink} and HO3D~\cite{hampali2021ho}.
To our knowledge, they are two of the few benchmarks that provide official contact annotations and corresponding RGB images.
OakInk is one of the latest and largest hand-object interaction datasets.
It contains 230K images, capturing the single-hand interactions of 12 subjects with 100 objects from 32 categories.
HO3D is a widely used dataset consisting of 103k images. The dataset captures 10 subjects interacting with 10 YCB objects~\cite{calli2015ycb}. 
More detailed dataset settings are provided in the supplementary material.

\subsection{Evaluation Metrics}
For contact prediction, detection metrics such as precision, recall, and F1-score are adopted.
For object reconstruction, the chamfer distance (CD, $mm$), F-score at 5mm and 10mm thresholds are reported.
To evaluate the quality of the relation between objects and hands, the 
penetration depth (PD, $cm$) and intersection volume (IV, $cm^{3}$) are computed.

\subsection{Experimental Results for Contact Prediction}
Since there is no specific method focused on predicting hand contact regions from monocular images, we first conduct ablation experiments on model settings, then compare and validate the effectiveness of the multi-level graphormer. Finally, we evaluate different levels of contact prediction.

\begin{table}[t]
\setlength\tabcolsep{5pt}
\centering
\begin{tabular}{lcccccc}
\toprule
Method & $\mathcal{L}_{vs}$ & OPC & WL & Precision & Recall & F1\\
\midrule
$M_{1}$ & \XSolidBrush & \XSolidBrush & \XSolidBrush & 0.270 & 0.176 & 0.189 \\
$M_{2}$ & \CheckmarkBold & \XSolidBrush & \XSolidBrush & 0.285 & 0.196 & 0.210 \\
$M_{3}$ & \CheckmarkBold & \CheckmarkBold & \XSolidBrush & 0.309 & 0.192 & 0.213 \\
$M_{4}$ & \CheckmarkBold & \CheckmarkBold & \CheckmarkBold & \textbf{0.332} & \textbf{0.245} & \textbf{0.262} \\
\bottomrule
\end{tabular}
\caption{Ablation study for vertex-level contact predictions on the OakInk dataset. From left to right are whether to use $\mathcal{L}_{vertex\_sub}$ ($\mathcal{L}_{vs}$), whether to only predict contact (OPC, otherwise reconstruct the hand mesh at the same time), and whether to use the weighted loss (WL).}
\label{contactablation}
\end{table}

\paragraph{Ablation Study.}
Table~\ref{contactablation} illustrates the quantitative ablation results for vertex-level contact predictions on the OakInk dataset. 
$M_{1}$ is designed to estimate the hand mesh and vertex-level contact at the same time.
It yields the overall lowest detection scores.
Compared with $M_{1}$, $M_{2}$ further uses the loss $\mathcal{L}_{vertex\_sub}$ calculated on the sub-mesh proposed in Training Details.
The precision, recall, and F1-score are improved by 5.6\%, 11.4\%, and 11.1\%, respectively, proving the effectiveness of multi-scale features aggregation based on the hand model in this task.
Different from $M_{2}$, $M_{3}$ does not reconstruct the hand mesh and only performs hand contact prediction. Although the recall drops slightly, its precision improves by 8.4\%, showing that focusing on a single task could make the network learn more effectively.
Finally, compared with $M_{3}$, $M_{4}$ uses weight priors for BCE losses in Equation~\ref{sdf_formula} and achieves a huge boost on all metrics (\emph{e.g.}, 27.6\% on recall and 23.0\% on F1), showing that the weight priors of contacts introduced in Training Details can provide useful guidance for the model.

\begin{table}[t]
\setlength\tabcolsep{3.2pt}
\centering
\begin{tabular}{lcccccc}
\toprule
\multirow{2}*{Method} & \multicolumn{3}{c}{OakInk} & \multicolumn{3}{c}{HO3D}\\
\cmidrule(r){2-7}
& P & R & F1 & P & R & F1 \\
\midrule
Single-Vertex & 0.332 & \textbf{0.245} & 0.262 & 0.476 & 0.422 & 0.416 \\
\textbf{Multi-Vertex} & \textbf{0.342} & 0.244 & \textbf{0.262} & \textbf{0.510} & \textbf{0.441} & \textbf{0.436} \\
\midrule
Single-Part & 0.770 & 0.753 & 0.728 & 0.710 & 0.723 & 0.672 \\
\textbf{Multi-Part} & \textbf{0.790} & \textbf{0.767} & \textbf{0.747} & \textbf{0.722} & \textbf{0.741} & \textbf{0.685} \\
\bottomrule
\end{tabular}
\caption{Comparison of different network architectures on OakInk and HO3D datasets. `P' is precision and `R' is recall.}
\label{multilevel}
\end{table}

\begin{figure}[t]
\centering
\includegraphics[width=1.0\columnwidth]{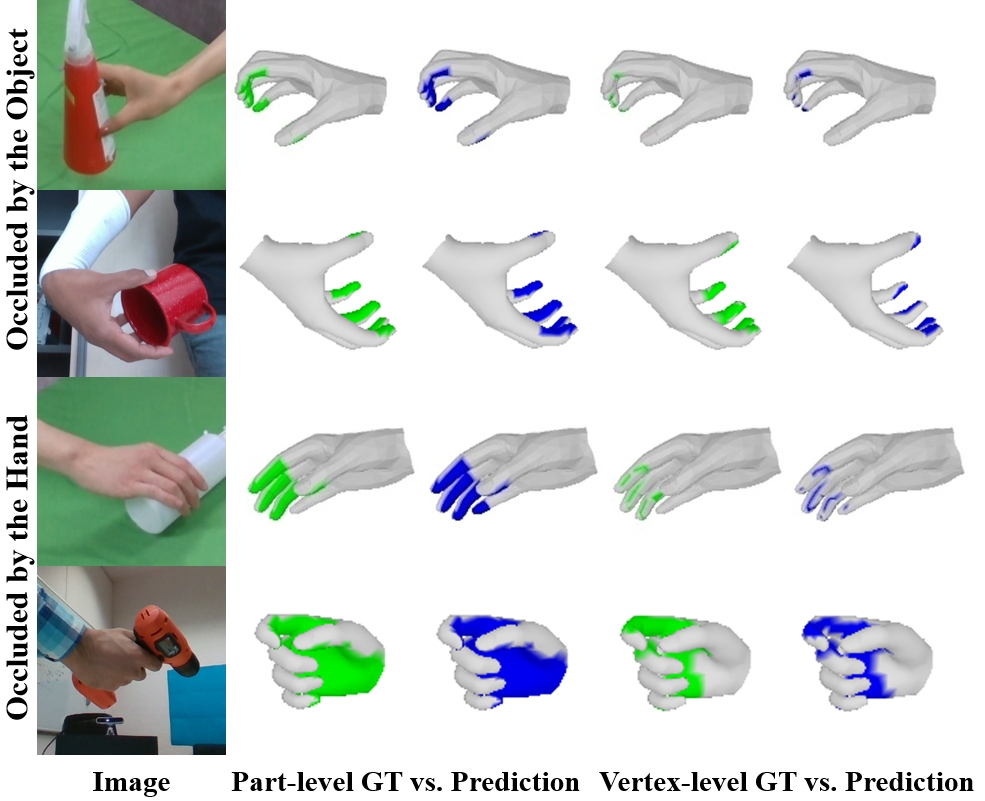}
\caption{Visualizations of contact prediction on OakInk (Rows 1, 3) and HO3D (Rows 2, 4) datasets.
Since the method only estimates contact, the result is rendered on the ground truth hand mesh. For samples whose contact regions are occluded by hands, hand meshes are rotated 180 degrees for clear visualization. The proposed method is robust to both hand and object occlusions.}
\label{contactVis}
\end{figure}

\paragraph{Effectiveness of Multi-level Graphormer.}
In this work, three network architectures are trained and evaluated on OakInk and HO3D benchmarks, respectively. 
As shown in Table~\ref{multilevel}, in addition to the multi-level graphormer encoders, we also use the single-level model in Fig.~\ref{pipeline} for contact prediction.
The outputs of the multi-level method are compared with corresponding single-level outputs. Although a single vertex-level model yields slightly better recall on OakInk, the multi-level one can improve the precision from 0.332 to 0.342 benefiting from using the part-level output to refine features for vertices. Regarding the part-level output, the coarse-to-fine model outperforms the single-level one on all evaluation metrics. The multi-level model also achieves superior performance on HO3D for all metrics, which demonstrates the advantage of using the proposed coarse-to-fine learning framework.
Fig.~\ref{contactVis} further illustrates the qualitative results of the proposed method on two benchmarks.
Benefiting from accumulating both global contexts and local details by using the graph-based transformer, the proposed method is robust to input images with hand or object occlusions.

\paragraph{Part-level vs. Vertex-level Prediction.}
In Table~\ref{multilevel}, the vertex-level predictions are worse than the part-level results, showing that the dense vertex-level prediction is more difficult than the sparse one.
For the single-level architecture, the F1 score of the vertex-level method on OakInk is only 0.262, while the single part-level model achieves 0.728.
On the HO3D dataset, we can observe a similar performance gap between the part-level and vertex-level accuracy. 
Fig.~\ref{contactVis} illustrates that part-level predictions are closer to the ground truth than vertex-level predictions.
Therefore, part-level predictions are converted to vertex-level ones according to the fixed correspondence and then propagated to hand mesh vertices for subsequent experiments. More details and comparisons are provided in the appendix.

\begin{table}[t]
\setlength\tabcolsep{2.8pt}
\centering
\begin{tabular}{lcccccc}
\toprule
Method & SPC & ESC & MSV & F@5mm$\uparrow$ & F@10mm$\uparrow$ & CD$\downarrow$ \\
\midrule
$N_{1}$ & \XSolidBrush & \CheckmarkBold & \CheckmarkBold & 0.261 & 0.475 & 1.110 \\
$N_{2}$ & \CheckmarkBold & \XSolidBrush & \CheckmarkBold & 0.361 & 0.592 & 0.848 \\
$N_{3}$ & \CheckmarkBold & \CheckmarkBold & \XSolidBrush & 0.371 & 0.614 & 0.680 \\
$N_{4}$ & \CheckmarkBold & \CheckmarkBold & \CheckmarkBold & \textbf{0.393} & \textbf{0.633} & \textbf{0.646} \\
\bottomrule
\end{tabular}
\caption{Ablation study for contact modeling in object reconstruction on HO3D. From left to right are whether to use the sparse convolution (SPC, otherwise nearest neighbor diffusion), whether to use the estimated contact (ESC, otherwise an all-one contact vector), and whether to use multi-scale contact code volumes (MSV, otherwise only $V_{3} + V_{4}$).}
\label{reconablation}
\end{table}

\subsection{Experimental Results on Object Reconstruction}

\paragraph{Ablation Study.}
Table~\ref{reconablation} illustrates the ablations for contact modeling in object reconstruction on HO3D and $N_{4}$ is our final method.
The ground truth hand meshes are adopted to ignore the influence of the hand pose.
$N_{1}$ is designed to diffuse the contact information by using a simple nearest neighbor method like LoopReg~\cite{bhatnagar2020loopreg}. It is slower and much worse than $N_{4}$ since the contact information is not fully learned like sparse convolution, which may even bring negative effects.
$N_{3}$ only combines two contact code volumes (i.e., $V_{3} + V_{4}$) and performs worse than $N_{4}$ with four scales, indicating the effectiveness of the full multi-scale contact code volumes.
In addition, $N_{2}$ removes the contact estimation module and takes an all-one contact vector for a fair comparison. Its results (e.g., $\rm {F@5mm}=0.361$) are worse than that of $N_{4}$ (e.g., $\rm {F@5mm}=0.393$), showing that the estimated contacts could provide flexible and efficient guidance for object reconstruction.
More ablation results can be found in the supplementary material.

\begin{table}[!t]
\centering
\begin{tabular}{lcccccc}
\toprule
Method & F@5mm & F@10mm & CD & PD$\downarrow$ & IV$\downarrow$ \\
\midrule
HO  & 0.110 & 0.220	& 4.190 & - & - \\
GF  & 0.120 & 0.240 & 4.960 & - & - \\
IHOI  & 0.280  & 0.500 & 1.530 & - & - \\
\textbf{Ours} & \textbf{0.313} & \textbf{0.542} & \textbf{1.081} & \textbf{1.02} & \textbf{5.11}  \\
\midrule
IHOI* & 0.351 & 0.600 & 0.656 & 0.90 & 4.10 \\
\textbf{Ours}* & \textbf{0.393} & \textbf{0.633} & \textbf{0.646} & \textbf{0.67} & \textbf{2.91}\\
\bottomrule
\end{tabular}
\caption{Comparison with state-of-the-art methods on HO3D. `*' denotes using the ground truth hand mesh.}
\label{ho3dResult}
\end{table}

\begin{table}[!t]
\setlength\tabcolsep{3.5pt}
\centering
\begin{tabular}{lcccccc}
\toprule
Method & F@5mm & F@10mm & CD & PD & IV \\
\midrule
IHOI & 0.432 & 0.658 & 0.491 & 0.75 & 4.36 \\
Ours w/o $\mathcal{L}_{hoi}$ & 0.447 & 0.716 & 0.274 & 0.66 & 3.03 \\
\textbf{Ours} & \textbf{0.459} & \textbf{0.718} & \textbf{0.260} & \textbf{0.62} & \textbf{2.67} \\
\bottomrule
\end{tabular}
\caption{Comparison on the OakInk benchmark.}
\label{oakinkResult}
\end{table}

\paragraph{Quantitative Comparison on HO3D.}
Since most prior methods require the object template during inference, the methods most relevant to ours are HO~\cite{hasson2019learning}, GF~\cite{karunratanakul2020grasping}, and IHOI~\cite{ye2022s}. 
For a fair comparison, we use the same predicted hands from~\cite{rong2020frankmocap} as IHOI and the estimated contact states from our multi-level model.
As shown in Table~\ref{ho3dResult}, when our model uses predicted hand meshes, we observe that our method can improve $\rm {F@5mm}$ and $\rm {F@10mm}$ by 11.8\% and 8.4\% and greatly reduce the chamfer distance by 29.3\%.
When the hand mesh is perfect, our method also shows an obvious advantage and consistently outperforms IHOI across all metrics. It largely improves $\rm {F@5mm}$ by 12.0\% and reduces the intersection volume by 29.0\%, demonstrating the superiority of our method in model-free object reconstruction.

\paragraph{Quantitative Comparison on OakInk.}
To show that our model can work well for unseen objects, we split the dataset to make sure that testing objects do not exist in the training set.
As shown in Table~\ref{oakinkResult}, when our model is not trained together with $L_{hoi}$, it can still outperform IHOI on all metrics. Our final model, which is learned with $L_{hoi}$, achieves even better results on different metrics. Compared with IHOI, our method can largely improve $\rm {F@5mm}$ and $\rm {F@10mm}$ by 6.3\% and 9.1\%, respectively. At the same time, it reduces the penetration depth and intersection volume by 17.3\% and 38.8\%, which suggests our model can reconstruct more realistic objects that naturally interact with hands.

\begin{figure}[t]
\centering
\includegraphics[width=1.0\columnwidth]{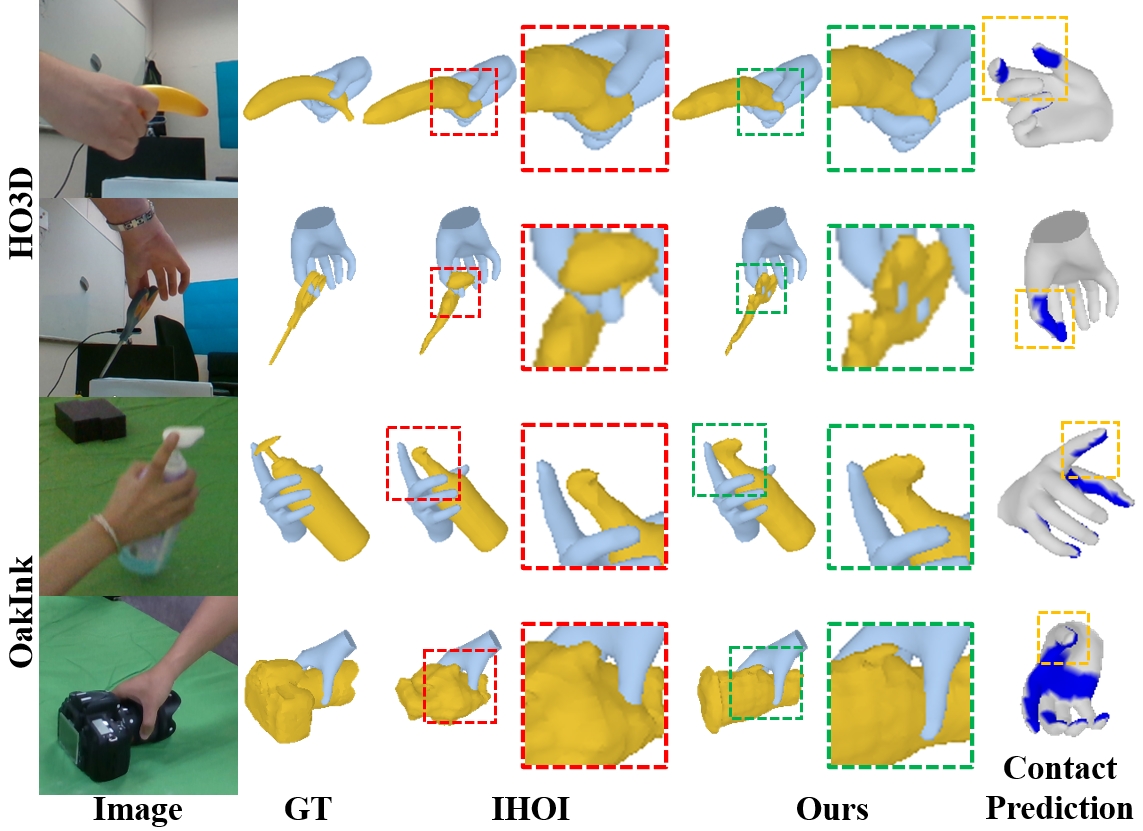}
\caption{Qualitative comparison with the state-of-the-art method on the HO3D and OakInk datasets. Our method can reconstruct more realistic objects, especially for parts that are in contact with hands.} 
\label{reconstructVis}
\end{figure}

\paragraph{Qualitative Comparison.}
Fig.~\ref{reconstructVis} illustrates qualitative comparisons on the HO3D and OakInk datasets.
Compared with the state-of-the-art IHOI~\cite{ye2022s} (red dotted box), our method shows a clear advantage in the reconstruction of object parts that are in contact with the hand. 
It can be seen that the predicted hand contacts (yellow dashed box) provide effective guidance to recover corresponding object parts (green dashed box).
We also observe that our method is robust to occlusions. As illustrated in the first, second, and fourth rows in Fig.~\ref{reconstructVis}, our model can still work well when objects are occluded by hands. For unseen objects with complex structures (\emph{e.g.}, camera) in OakInk, our model can also obtain realistic results. More qualitative results can be found in the supplementary material.


\section{Conclusion}

This paper introduces a novel representation of explicit contacts for implicit reconstruction of hand-held objects.
First, the multi-level graph-based transformer encoders are cascaded to  estimate accurate 3D hand-object contacts from a single RGB image.
Then, the predicted contact states are anchored to the hand surface and diffused to the nearby space to construct the implicit neural representation for the manipulated object. 
Extensive experiments on HO3D and OakInk datasets indicate that our method can pay more attention to the object parts that are in contact with hands and reconstruct more realistic object meshes.
The proposed method currently focuses on hand-held object reconstruction.
In future work, we attempt to integrate the hand reconstruction module for better hand-object interaction reconstruction and leverage object category priors to improve generalization.

\section{Acknowledgments}
This work is supported by the National Natural Science Foundation of China under Grant No. U23B2054, 62006225, 62071468, and also funded by the National Key Research and Development Program of China under Grant No. 2021ZD0113503 and 2022YFC3310400, the Fundamental Research Funds for the Central Universities under Grant No. 2233100028.
We would like to thank Liang An and Yuxiang Zhang for their help, feedback, and discussions in the early work of this paper.

\bibliography{aaai24}

\begin{thebibliography}{67}
\providecommand{\natexlab}[1]{#1}

\bibitem[{Baek, Kim, and Kim(2019)}]{baek2019pushing}
Baek, S.; Kim, K.~I.; and Kim, T.-K. 2019.
\newblock Pushing the envelope for {RGB}-based dense {3D} hand pose estimation
  via neural rendering.
\newblock In \emph{CVPR}.

\bibitem[{Bhatnagar et~al.(2020)Bhatnagar, Sminchisescu, Theobalt, and
  Pons-Moll}]{bhatnagar2020loopreg}
Bhatnagar, B.~L.; Sminchisescu, C.; Theobalt, C.; and Pons-Moll, G. 2020.
\newblock Loopreg: Self-supervised learning of implicit surface
  correspondences, pose and shape for 3d human mesh registration.
\newblock \emph{NeurIPS}.

\bibitem[{Bicchi and Kumar(2000)}]{bicchi2000robotic}
Bicchi, A.; and Kumar, V. 2000.
\newblock Robotic grasping and contact: A review.
\newblock In \emph{ICRA}.

\bibitem[{Boukhayma, Bem, and Torr(2019)}]{boukhayma20193d}
Boukhayma, A.; Bem, R.~d.; and Torr, P.~H. 2019.
\newblock {3D} hand shape and pose from images in the wild.
\newblock In \emph{CVPR}.

\bibitem[{Buescher et~al.(2015)Buescher, Meier, Walck, Haschke, and
  Ritter}]{buescher2015augmenting}
Buescher, G.; Meier, M.; Walck, G.; Haschke, R.; and Ritter, H.~J. 2015.
\newblock Augmenting curved robot surfaces with soft tactile skin.
\newblock In \emph{IROS}.

\bibitem[{Calli et~al.(2015)Calli, Singh, Walsman, Srinivasa, Abbeel, and
  Dollar}]{calli2015ycb}
Calli, B.; Singh, A.; Walsman, A.; Srinivasa, S.; Abbeel, P.; and Dollar, A.~M.
  2015.
\newblock The ycb object and model set: Towards common benchmarks for
  manipulation research.
\newblock In \emph{ICAR}.

\bibitem[{Chao et~al.(2021)Chao, Yang, Xiang, Molchanov, Handa, Tremblay,
  Narang, Van~Wyk, Iqbal, Birchfield et~al.}]{chao2021dexycb}
Chao, Y.-W.; Yang, W.; Xiang, Y.; Molchanov, P.; Handa, A.; Tremblay, J.;
  Narang, Y.~S.; Van~Wyk, K.; Iqbal, U.; Birchfield, S.; et~al. 2021.
\newblock {DexYCB}: A benchmark for capturing hand grasping of objects.
\newblock In \emph{CVPR}.

\bibitem[{Chen et~al.(2021)Chen, Liu, Ma, Chang, Wang, Chen, Guo, Wan, and
  Zheng}]{chen2021camera}
Chen, X.; Liu, Y.; Ma, C.; Chang, J.; Wang, H.; Chen, T.; Guo, X.; Wan, P.; and
  Zheng, W. 2021.
\newblock Camera-space hand mesh recovery via semantic aggregation and adaptive
  {2D-1D} registration.
\newblock In \emph{CVPR}.

\bibitem[{Chen et~al.(2023{\natexlab{a}})Chen, Dwivedi, Black, and
  Tzionas}]{chen2023detecting}
Chen, Y.; Dwivedi, S.~K.; Black, M.~J.; and Tzionas, D. 2023{\natexlab{a}}.
\newblock Detecting Human-Object Contact in Images.
\newblock In \emph{CVPR}.

\bibitem[{Chen et~al.(2023{\natexlab{b}})Chen, Chen, Schmid, and
  Laptev}]{chen2023gsdf}
Chen, Z.; Chen, S.; Schmid, C.; and Laptev, I. 2023{\natexlab{b}}.
\newblock {gSDF}: {Geometry-Driven} Signed Distance Functions for {3D}
  Hand-Object Reconstruction.
\newblock In \emph{CVPR}.

\bibitem[{Chen et~al.(2022)Chen, Hasson, Schmid, and Laptev}]{chen2022alignsdf}
Chen, Z.; Hasson, Y.; Schmid, C.; and Laptev, I. 2022.
\newblock {AlignSDF}: Pose-Aligned Signed Distance Fields for Hand-Object
  Reconstruction.
\newblock In \emph{ECCV}.

\bibitem[{Chen and Zhang(2019)}]{chen2019learning}
Chen, Z.; and Zhang, H. 2019.
\newblock Learning implicit fields for generative shape modeling.
\newblock In \emph{CVPR}.

\bibitem[{Choi et~al.(2022)Choi, Moon, Armando, Leroy, Lee, and
  Rogez}]{choi2022mononhr}
Choi, H.; Moon, G.; Armando, M.; Leroy, V.; Lee, K.~M.; and Rogez, G. 2022.
\newblock Mononhr: Monocular neural human renderer.
\newblock In \emph{3DV}.

\bibitem[{Choy et~al.(2016)Choy, Xu, Gwak, Chen, and Savarese}]{choy20163d}
Choy, C.~B.; Xu, D.; Gwak, J.; Chen, K.; and Savarese, S. 2016.
\newblock {3D-R2N2}: A unified approach for single and multi-view {3D} object
  reconstruction.
\newblock In \emph{ECCV}.

\bibitem[{Fieraru et~al.(2020)Fieraru, Zanfir, Oneata, Popa, Olaru, and
  Sminchisescu}]{fieraru2020three}
Fieraru, M.; Zanfir, M.; Oneata, E.; Popa, A.-I.; Olaru, V.; and Sminchisescu,
  C. 2020.
\newblock Three-dimensional reconstruction of human interactions.
\newblock In \emph{CVPR}.

\bibitem[{Fieraru et~al.(2021)Fieraru, Zanfir, Oneata, Popa, Olaru, and
  Sminchisescu}]{fieraru2021learning}
Fieraru, M.; Zanfir, M.; Oneata, E.; Popa, A.-I.; Olaru, V.; and Sminchisescu,
  C. 2021.
\newblock Learning complex 3D human self-contact.
\newblock In \emph{AAAI}.

\bibitem[{Grady et~al.(2021)Grady, Tang, Twigg, Vo, Brahmbhatt, and
  Kemp}]{grady2021contactopt}
Grady, P.; Tang, C.; Twigg, C.~D.; Vo, M.; Brahmbhatt, S.; and Kemp, C.~C.
  2021.
\newblock {ContactOpt}: Optimizing contact to improve grasps.
\newblock In \emph{CVPR}.

\bibitem[{Graham, Engelcke, and Van Der~Maaten(2018)}]{graham20183d}
Graham, B.; Engelcke, M.; and Van Der~Maaten, L. 2018.
\newblock {3D} semantic segmentation with submanifold sparse convolutional
  networks.
\newblock In \emph{CVPR}.

\bibitem[{Groueix et~al.(2018)Groueix, Fisher, Kim, Russell, and
  Aubry}]{groueix2018papier}
Groueix, T.; Fisher, M.; Kim, V.~G.; Russell, B.~C.; and Aubry, M. 2018.
\newblock A papier-m{\^a}ch{\'e} approach to learning 3d surface generation.
\newblock In \emph{CVPR}.

\bibitem[{Hampali et~al.(2020)Hampali, Rad, Oberweger, and
  Lepetit}]{hampali2020honnotate}
Hampali, S.; Rad, M.; Oberweger, M.; and Lepetit, V. 2020.
\newblock {HOnnotate}: A method for {3D} annotation of hand and object poses.
\newblock In \emph{CVPR}.

\bibitem[{Hampali, Sarkar, and Lepetit(2021)}]{hampali2021ho}
Hampali, S.; Sarkar, S.~D.; and Lepetit, V. 2021.
\newblock HO-3D\_v3: Improving the accuracy of hand-object annotations of the
  {HO-3D} dataset.
\newblock \emph{arXiv preprint arXiv:2107.00887}.

\bibitem[{Hampali et~al.(2022)Hampali, Sarkar, Rad, and
  Lepetit}]{hampali2022keypoint}
Hampali, S.; Sarkar, S.~D.; Rad, M.; and Lepetit, V. 2022.
\newblock {Keypoint Transformer}: Solving Joint Identification in Challenging
  Hands and Object Interactions for Accurate {3D} Pose Estimation.
\newblock In \emph{CVPR}.

\bibitem[{Hasson et~al.(2020)Hasson, Tekin, Bogo, Laptev, Pollefeys, and
  Schmid}]{hasson2020leveraging}
Hasson, Y.; Tekin, B.; Bogo, F.; Laptev, I.; Pollefeys, M.; and Schmid, C.
  2020.
\newblock Leveraging photometric consistency over time for sparsely supervised
  hand-object reconstruction.
\newblock In \emph{CVPR}.

\bibitem[{Hasson et~al.(2019)Hasson, Varol, Tzionas, Kalevatykh, Black, Laptev,
  and Schmid}]{hasson2019learning}
Hasson, Y.; Varol, G.; Tzionas, D.; Kalevatykh, I.; Black, M.~J.; Laptev, I.;
  and Schmid, C. 2019.
\newblock Learning joint reconstruction of hands and manipulated objects.
\newblock In \emph{CVPR}.

\bibitem[{Hu et~al.(2022)Hu, Yi, Zhang, Yong, and Xu}]{hu2022physical}
Hu, H.; Yi, X.; Zhang, H.; Yong, J.-H.; and Xu, F. 2022.
\newblock Physical Interaction: Reconstructing Hand-object Interactions with
  Physics.
\newblock In \emph{SIGGRAPH Asia}.

\bibitem[{Huang et~al.(2022)Huang, Yi, H{\"o}schle, Safroshkin, Alexiadis,
  Polikovsky, Scharstein, and Black}]{huang2022capturing}
Huang, C.-H.~P.; Yi, H.; H{\"o}schle, M.; Safroshkin, M.; Alexiadis, T.;
  Polikovsky, S.; Scharstein, D.; and Black, M.~J. 2022.
\newblock Capturing and inferring dense full-body human-scene contact.
\newblock In \emph{CVPR}.

\bibitem[{Jain et~al.(2019)Jain, Li, Singhal, Rajeswaran, Kumar, and
  Todorov}]{jain2019learning}
Jain, D.; Li, A.; Singhal, S.; Rajeswaran, A.; Kumar, V.; and Todorov, E. 2019.
\newblock Learning deep visuomotor policies for dexterous hand manipulation.
\newblock In \emph{ICRA}.

\bibitem[{Karunratanakul et~al.(2020)Karunratanakul, Yang, Zhang, Black,
  Muandet, and Tang}]{karunratanakul2020grasping}
Karunratanakul, K.; Yang, J.; Zhang, Y.; Black, M.~J.; Muandet, K.; and Tang,
  S. 2020.
\newblock {Grasping Field}: Learning implicit representations for human grasps.
\newblock In \emph{3DV}.

\bibitem[{Kingma and Ba(2015)}]{KingmaB14}
Kingma, D.~P.; and Ba, J. 2015.
\newblock Adam: {A} Method for Stochastic Optimization.
\newblock In \emph{ICLR}.

\bibitem[{Kipf and Welling(2017)}]{kipf2017semi}
Kipf, T.~N.; and Welling, M. 2017.
\newblock Semi-supervised classification with graph convolutional networks.
\newblock In \emph{ICLR}.

\bibitem[{Kocabas et~al.(2021)Kocabas, Huang, Hilliges, and
  Black}]{kocabas2021pare}
Kocabas, M.; Huang, C.-H.~P.; Hilliges, O.; and Black, M.~J. 2021.
\newblock PARE: Part attention regressor for 3D human body estimation.
\newblock In \emph{ICCV}.

\bibitem[{Kulon et~al.(2020)Kulon, G{\"u}ler, Kokkinos, Bronstein, and
  Zafeiriou}]{Kulon2020weaklysupervisedmh}
Kulon, D.; G{\"u}ler, R.~A.; Kokkinos, I.; Bronstein, M.; and Zafeiriou, S.
  2020.
\newblock Weakly-Supervised Mesh-Convolutional Hand Reconstruction in the Wild.
\newblock In \emph{CVPR}.

\bibitem[{Kwon et~al.(2021)Kwon, Kim, Ceylan, and Fuchs}]{kwon2021neural}
Kwon, Y.; Kim, D.; Ceylan, D.; and Fuchs, H. 2021.
\newblock Neural human performer: Learning generalizable radiance fields for
  human performance rendering.
\newblock \emph{NeurIPS}.

\bibitem[{Kyriazis and Argyros(2014)}]{kyriazis2014scalable}
Kyriazis, N.; and Argyros, A. 2014.
\newblock Scalable {3D} tracking of multiple interacting objects.
\newblock In \emph{CVPR}.

\bibitem[{Li et~al.(2022)Li, An, Zhang, Wu, Chen, Yu, and
  Liu}]{li2022interacting}
Li, M.; An, L.; Zhang, H.; Wu, L.; Chen, F.; Yu, T.; and Liu, Y. 2022.
\newblock Interacting attention graph for single image two-hand reconstruction.
\newblock In \emph{CVPR}.

\bibitem[{Li et~al.(2020)Li, Kroemer, Su, Veiga, Kaboli, and
  Ritter}]{li2020review}
Li, Q.; Kroemer, O.; Su, Z.; Veiga, F.~F.; Kaboli, M.; and Ritter, H.~J. 2020.
\newblock A review of tactile information: Perception and action through touch.
\newblock \emph{TRO}.

\bibitem[{Lin, Wang, and Liu(2021)}]{lin2021mesh}
Lin, K.; Wang, L.; and Liu, Z. 2021.
\newblock Mesh graphormer.
\newblock In \emph{CVPR}.

\bibitem[{Mescheder et~al.(2019)Mescheder, Oechsle, Niemeyer, Nowozin, and
  Geiger}]{mescheder2019occupancy}
Mescheder, L.; Oechsle, M.; Niemeyer, M.; Nowozin, S.; and Geiger, A. 2019.
\newblock Occupancy {Networks}: Learning {3D} reconstruction in function space.
\newblock In \emph{CVPR}.

\bibitem[{Mildenhall et~al.(2022)Mildenhall, Srinivasan, Tancik, Barron,
  Ramamoorthi, and Ng}]{mildenhall2021nerf}
Mildenhall, B.; Srinivasan, P.~P.; Tancik, M.; Barron, J.~T.; Ramamoorthi, R.;
  and Ng, R. 2022.
\newblock {NeRF}: Representing scenes as neural radiance fields for view
  synthesis.
\newblock In \emph{ECCV}.

\bibitem[{Mundy(2006)}]{mundy2006geometric}
Mundy, J.~L. 2006.
\newblock Object Recognition in the Geometric Era: {A} Retrospective.
\newblock In \emph{Toward Category-Level Object Recognition}, Lecture Notes in
  Computer Science.

\bibitem[{Oikonomidis, Kyriazis, and Argyros(2011)}]{oikonomidis2011full}
Oikonomidis, I.; Kyriazis, N.; and Argyros, A.~A. 2011.
\newblock Full {DOF} tracking of a hand interacting with an object by modeling
  occlusions and physical constraints.
\newblock In \emph{ICCV}.

\bibitem[{Park et~al.(2019)Park, Florence, Straub, Newcombe, and
  Lovegrove}]{park2019deepsdf}
Park, J.~J.; Florence, P.; Straub, J.; Newcombe, R.; and Lovegrove, S. 2019.
\newblock {DeepSDF}: Learning continuous signed distance functions for shape
  representation.
\newblock In \emph{CVPR}.

\bibitem[{Paszke et~al.(2019)Paszke, Gross, Massa, Lerer, Bradbury, Chanan,
  Killeen, Lin, Gimelshein, Antiga et~al.}]{paszke2019pytorch}
Paszke, A.; Gross, S.; Massa, F.; Lerer, A.; Bradbury, J.; Chanan, G.; Killeen,
  T.; Lin, Z.; Gimelshein, N.; Antiga, L.; et~al. 2019.
\newblock Pytorch: An imperative style, high-performance deep learning library.
\newblock \emph{NeurIPS}.

\bibitem[{Pavlakos et~al.(2017)Pavlakos, Zhou, Derpanis, and
  Daniilidis}]{pavlakos2017coarse}
Pavlakos, G.; Zhou, X.; Derpanis, K.~G.; and Daniilidis, K. 2017.
\newblock Coarse-to-fine volumetric prediction for single-image {3D} human
  pose.
\newblock In \emph{CVPR}.

\bibitem[{Peng et~al.(2021{\natexlab{a}})Peng, Jiang, Liao, Niemeyer,
  Pollefeys, and Geiger}]{peng2021shape}
Peng, S.; Jiang, C.; Liao, Y.; Niemeyer, M.; Pollefeys, M.; and Geiger, A.
  2021{\natexlab{a}}.
\newblock Shape as points: A differentiable poisson solver.
\newblock \emph{NeurIPS}.

\bibitem[{Peng et~al.(2021{\natexlab{b}})Peng, Zhang, Xu, Wang, Shuai, Bao, and
  Zhou}]{peng2021neural}
Peng, S.; Zhang, Y.; Xu, Y.; Wang, Q.; Shuai, Q.; Bao, H.; and Zhou, X.
  2021{\natexlab{b}}.
\newblock Neural body: Implicit neural representations with structured latent
  codes for novel view synthesis of dynamic humans.
\newblock In \emph{CVPR}.

\bibitem[{Qi et~al.(2017{\natexlab{a}})Qi, Su, Mo, and Guibas}]{qi2017pointnet}
Qi, C.~R.; Su, H.; Mo, K.; and Guibas, L.~J. 2017{\natexlab{a}}.
\newblock {PointNet}: Deep learning on point sets for {3D} classification and
  segmentation.
\newblock In \emph{CVPR}.

\bibitem[{Qi et~al.(2017{\natexlab{b}})Qi, Yi, Su, and
  Guibas}]{qi2017pointnet++}
Qi, C.~R.; Yi, L.; Su, H.; and Guibas, L.~J. 2017{\natexlab{b}}.
\newblock {PointNet}++: Deep hierarchical feature learning on point sets in a
  metric space.
\newblock \emph{NeurIPS}.

\bibitem[{Riegler, Ulusoy, and Geiger(2017)}]{Riegler2017OctNet}
Riegler, G.; Ulusoy, A.~O.; and Geiger, A. 2017.
\newblock {OctNet}: Learning Deep {3D} Representations at High Resolutions.
\newblock In \emph{CVPR}.

\bibitem[{Roberts(1963)}]{roberts_thesis_1963}
Roberts, L.~G. 1963.
\newblock \emph{Machine perception of three-dimensional solids}.
\newblock Ph.D. thesis, Massachusetts Institute of Technology.

\bibitem[{Romero, Tzionas, and Black(2017)}]{MANO:SIGGRAPHASIA:2017}
Romero, J.; Tzionas, D.; and Black, M.~J. 2017.
\newblock Embodied {Hands}: Modeling and Capturing Hands and Bodies Together.
\newblock \emph{TOG}.

\bibitem[{Rong, Shiratori, and Joo(2020)}]{rong2020frankmocap}
Rong, Y.; Shiratori, T.; and Joo, H. 2020.
\newblock {FrankMocap}: Fast monocular {3D} hand and body motion capture by
  regression and integration.
\newblock \emph{arXiv preprint arXiv:2008.08324}.

\bibitem[{Tse et~al.(2022{\natexlab{a}})Tse, Kim, Leonardis, and
  Chang}]{tse2022collaborative}
Tse, T. H.~E.; Kim, K.~I.; Leonardis, A.; and Chang, H.~J. 2022{\natexlab{a}}.
\newblock Collaborative Learning for Hand and Object Reconstruction with
  Attention-guided Graph Convolution.
\newblock In \emph{CVPR}.

\bibitem[{Tse et~al.(2022{\natexlab{b}})Tse, Zhang, Kim, Leonardis, Zheng, and
  Chang}]{tse2022s}
Tse, T. H.~E.; Zhang, Z.; Kim, K.~I.; Leonardis, A.; Zheng, F.; and Chang,
  H.~J. 2022{\natexlab{b}}.
\newblock S2Contact: Graph-Based Network for 3D Hand-Object Contact Estimation
  with Semi-supervised Learning.
\newblock In \emph{ECCV}.

\bibitem[{Vaswani et~al.(2017)Vaswani, Shazeer, Parmar, Uszkoreit, Jones,
  Gomez, Kaiser, and Polosukhin}]{Vaswani2017Attention}
Vaswani, A.; Shazeer, N.; Parmar, N.; Uszkoreit, J.; Jones, L.; Gomez, A.~N.;
  Kaiser, L.~u.; and Polosukhin, I. 2017.
\newblock Attention is All you Need.
\newblock In \emph{NeurIPS}.

\bibitem[{Wang et~al.(2020)Wang, Sun, Cheng, Jiang, Deng, Zhao, Liu, Mu, Tan,
  Wang et~al.}]{wang2020deep}
Wang, J.; Sun, K.; Cheng, T.; Jiang, B.; Deng, C.; Zhao, Y.; Liu, D.; Mu, Y.;
  Tan, M.; Wang, X.; et~al. 2020.
\newblock Deep high-resolution representation learning for visual recognition.
\newblock \emph{TPAMI}.

\bibitem[{Wang et~al.(2018)Wang, Zhang, Li, Fu, Liu, and
  Jiang}]{wang2018pixel2mesh}
Wang, N.; Zhang, Y.; Li, Z.; Fu, Y.; Liu, W.; and Jiang, Y.-G. 2018.
\newblock Pixel2mesh: Generating {3D} mesh models from single {RGB} images.
\newblock In \emph{ECCV}.

\bibitem[{Wang et~al.(2013)Wang, Min, Zhang, Liu, Xu, Dai, and
  Chai}]{wang2013video}
Wang, Y.; Min, J.; Zhang, J.; Liu, Y.; Xu, F.; Dai, Q.; and Chai, J. 2013.
\newblock Video-based hand manipulation capture through composite motion
  control.
\newblock \emph{TOG}.

\bibitem[{Yang et~al.(2022{\natexlab{a}})Yang, Li, Zhan, Lv, Xu, Li, and
  Lu}]{yang2022artiboost}
Yang, L.; Li, K.; Zhan, X.; Lv, J.; Xu, W.; Li, J.; and Lu, C.
  2022{\natexlab{a}}.
\newblock {ArtiBoost}: Boosting Articulated {3D} Hand-Object Pose Estimation
  via Online Exploration and Synthesis.
\newblock In \emph{CVPR}.

\bibitem[{Yang et~al.(2022{\natexlab{b}})Yang, Li, Zhan, Wu, Xu, Liu, and
  Lu}]{yang2022oakink}
Yang, L.; Li, K.; Zhan, X.; Wu, F.; Xu, A.; Liu, L.; and Lu, C.
  2022{\natexlab{b}}.
\newblock {OakInk}: A Large-scale Knowledge Repository for Understanding
  Hand-Object Interaction.
\newblock In \emph{CVPR}.

\bibitem[{Yang et~al.(2021)Yang, Zhan, Li, Xu, Li, and Lu}]{yang2021cpf}
Yang, L.; Zhan, X.; Li, K.; Xu, W.; Li, J.; and Lu, C. 2021.
\newblock {CPF}: Learning a contact potential field to model the hand-object
  interaction.
\newblock In \emph{ICCV}.

\bibitem[{Ye, Gupta, and Tulsiani(2022)}]{ye2022s}
Ye, Y.; Gupta, A.; and Tulsiani, S. 2022.
\newblock What's in your hands? {3D} Reconstruction of Generic Objects in
  Hands.
\newblock In \emph{CVPR}.

\bibitem[{Yin et~al.(2023)Yin, Huang, Qin, Chen, and Wang}]{touch-dexterity}
Yin, Z.-H.; Huang, B.; Qin, Y.; Chen, Q.; and Wang, X. 2023.
\newblock Rotating without Seeing: Towards In-hand Dexterity through Touch.
\newblock \emph{RSS}.

\bibitem[{Zhang et~al.(2020)Zhang, Cao, Lu, Ouyang, and
  Sun}]{zhang2020learning}
Zhang, H.; Cao, J.; Lu, G.; Ouyang, W.; and Sun, Z. 2020.
\newblock Learning 3d human shape and pose from dense body parts.
\newblock \emph{TPAMI}, 44(5): 2610--2627.

\bibitem[{Zhang et~al.(2023)Zhang, Tian, Zhang, Li, An, Sun, and
  Liu}]{zhang2023pymaf}
Zhang, H.; Tian, Y.; Zhang, Y.; Li, M.; An, L.; Sun, Z.; and Liu, Y. 2023.
\newblock Pymaf-x: Towards well-aligned full-body model regression from
  monocular images.
\newblock \emph{TPAMI}, 45(10): 12287--12303.

\bibitem[{Zhang et~al.(2021)Zhang, Zhou, Tian, Yong, and Xu}]{zhang2021single}
Zhang, H.; Zhou, Y.; Tian, Y.; Yong, J.-H.; and Xu, F. 2021.
\newblock Single depth view based real-time reconstruction of hand-object
  interactions.
\newblock \emph{TOG}.

\bibitem[{Zhao et~al.(2022)Zhao, Zuo, Xie, and Wang}]{zhao2022stability}
Zhao, Z.; Zuo, B.; Xie, W.; and Wang, Y. 2022.
\newblock Stability-driven contact reconstruction from monocular color images.
\newblock In \emph{CVPR}.

\end{thebibliography}


\section*{\huge{Appendix}}
\appendix


\section{Dataset Details}\label{Dataset}
In this paper, two real-world datasets, OakInk~\cite{yang2022oakink} and HO3D~\cite{hampali2021ho} are used for comparison.

\begin{figure}[t]
\centering
\includegraphics[width=1.0\columnwidth]{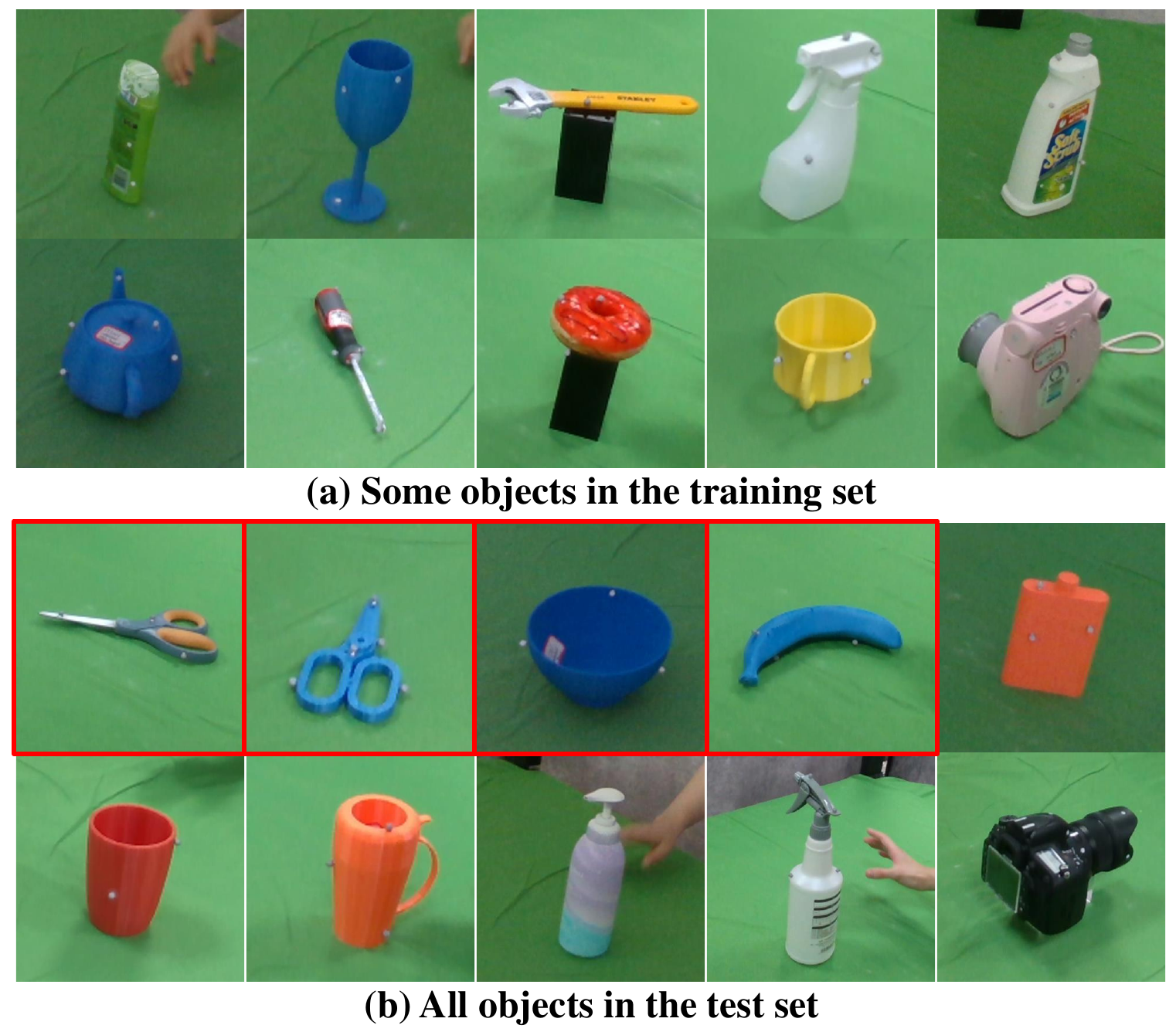}
\caption{Some training and all test objects in the OakInk dataset. Objects in the red box belong to categories unique to the test set.}
\label{oakinksamples}
\end{figure}

\paragraph{OakInk.}
The OakInk benchmark contains 100 objects from 32 categories. 
For contact prediction, the official split originally used for hand-object pose estimation~\cite{yang2022oakink} is adopted. The samples with a minimum distance between hand and object vertices greater than 5 mm are filtered out, resulting in 145,589 training images and 48,538 testing images.
For hand-held object reconstruction, to verify the generalization of the model, we randomly select 10 objects (i.e., A01027, A02031, C12001, C35001, C91001, O50001, S10008, S10018, S16003, Y35037) from the above testing set and mark all their images as a new testing set. The training set only keeps images of the other 90 objects. Finally, we obtain 131,287 samples for training and 4,773 for testing.
Fig.~\ref{oakinksamples} shows some training and all test objects in this dataset.

\paragraph{HO3D.}
HO3D consists of 10 subjects interacting with 10 YCB objects~\cite{calli2015ycb}. For a fair comparison with state-of-the-art methods~\cite{hasson2019learning,karunratanakul2020grasping,ye2022s} in Table 4 of the main text, we follow the data partition of IHOI~\cite{ye2022s} and use the predicted hand meshes from~\cite{rong2020frankmocap}. The estimated contact states utilized in the proposed method are produced by our multi-level model. 
In addition, some samples in HO3D v3~\cite{hampali2021ho} lack contact annotations, making it impossible to evaluate contact prediction methods on them. Therefore, we further remove samples without contact labels in HO3D and use the resulting subset for other experiments, resulting in 64,775 images for training and 1,032 images for testing. 

\begin{figure*}[t]
\centering
\includegraphics[width=1.0\textwidth]{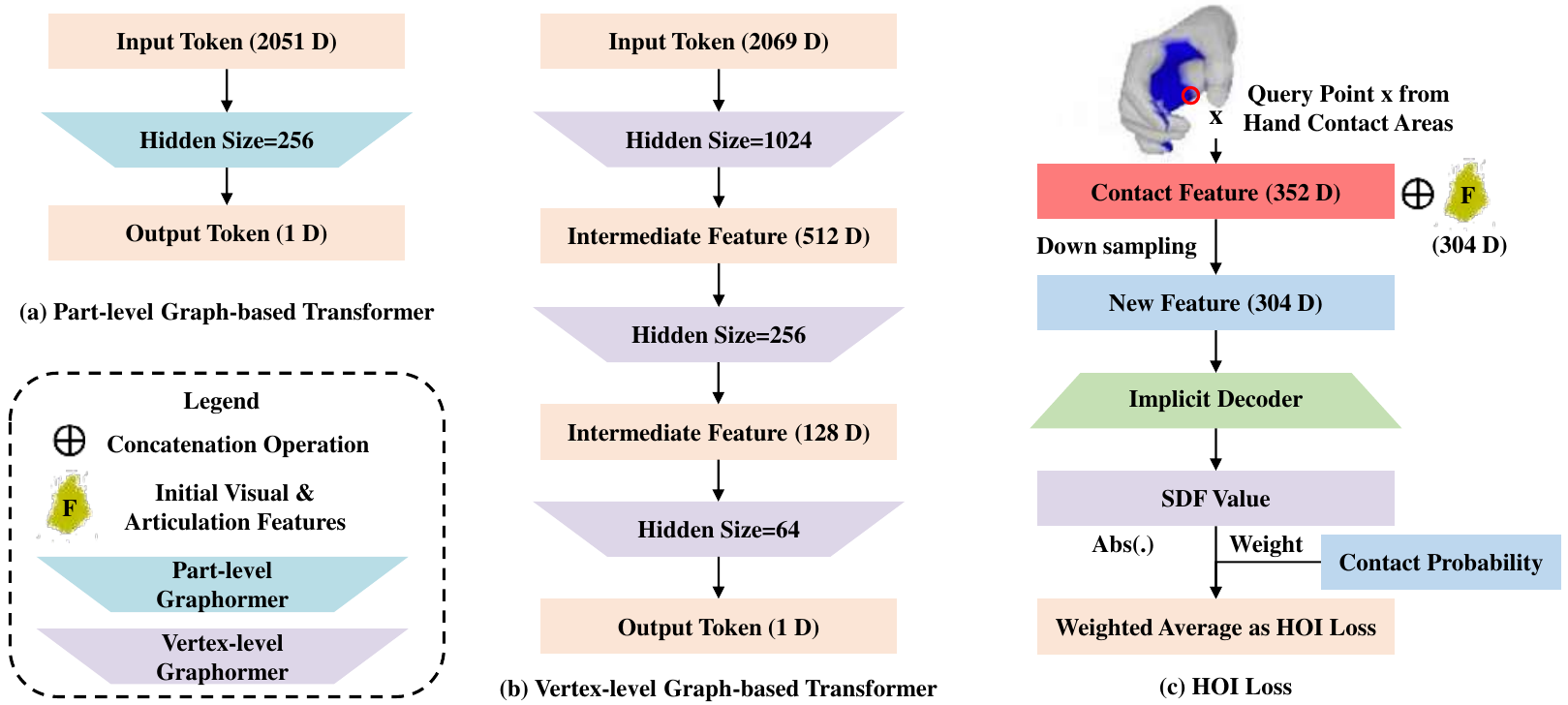}
\caption{Network Architecture.
(a) The architecture of the part-level graph-based transformer.
(b) The architecture of the vertex-level graph-based transformer.
(c) The computation process of $\mathcal{L}_{hoi}$.}
\label{networks}
\end{figure*}

\begin{table*}[ht]
\scriptsize
\centering
\caption{Comparison of different methods for contact prediction and corresponding object reconstruction on the HO3D benchmark. The value in brackets indicates the difference between the current and the previous row. For object reconstruction metrics, we provide the mean/standard deviation. The ground truth hand meshes are adopted to ignore the influence of the hand pose.}
\label{contact_compare}
\begin{tabular}{lcccccc}
\toprule
Method & Precision$\uparrow$ & Recall$\uparrow$ & F1$\uparrow$ & F@5mm$\uparrow$ & F@10mm$\uparrow$ & Chamfer Distance ($mm$)$\downarrow$ \\
\midrule
BSTRO   & 0.467  & 0.416 &	0.400 &	0.363/0.007 & 0.607/0.010 & 0.764/0.035 \\
Single Vertex-level  & 0.476 (1.9\%$\uparrow$) & 0.422 (1.4\%$\uparrow$) & 0.416 (4.0\%$\uparrow$)& 0.371 (2.2\% $\uparrow$)/0.005 & 0.615 (1.3\%$\uparrow$)/0.007 & 0.739 (3.3\%$\downarrow$)/0.013 \\
Multi-level (Vertex output) & \textbf{0.510} (7.1\% $\uparrow$) & \textbf{0.441} (4.5\%$\uparrow$) & \textbf{0.436} (4.8\% $\uparrow$) & \textbf{0.374} (0.8\% $\uparrow$)/0.004 & \textbf{0.620} (0.8\%$\uparrow$)/0.006 & \textbf{0.701} (5.1\%$\downarrow$)/0.059 \\
\bottomrule
\end{tabular}
\end{table*}

\section{Network Architecture}\label{Network}
In Fig.~\ref{networks}, we illustrate more details about our network architecture. The proposed graph-based transformer consists of graphormer blocks proposed in~\cite{lin2021mesh}. 
For the part-level graph-based transformer (Fig.~\ref{networks} (a)), one block is sufficient while three blocks are utilized for the vertex-level estimator (Fig.~\ref{networks} (b)). The intermediate features and their dimensions can be found in the figure.
In addition, the computation of the hand-object interaction loss $\mathcal{L}_{hoi}$ is shown in Fig.~\ref{networks} (c).
The dimension of the extracted contact feature is 352, and the initial features $F$~\cite{ye2022s} including visual and articulation embeddings are 304 dimensions. A linear layer is used to downsample the concatenated features to 304 dimensions for a fair comparison with IHOI~\cite{ye2022s}.


\section{Quantitative Results}\label{Quantitative}

\subsection{Comparison with Contact Prediction Methods}
To our knowledge, estimating contact states from a single RGB image is mainly explored for human body mesh without focusing on the hand.
Therefore, we reimplement a version of hand contact estimation based on BSTRO~\cite{huang2022capturing}, which is designed for human contact prediction. Since BSTRO only estimates the vertex-level contact, it is similar to our single vertex-level method. Table~\ref{contact_compare} illustrates that the BSTRO-based method (e.g., F1=0.400) performs worse than our single vertex-level model (e.g., F1=0.416) and also worse than our multi-level model (e.g., F1=0.436). 
Although both are based on transformer architecture, there are three main differences between BSTRO and the proposed method. The first two differences are the distinct application scenarios and the multi-level framework. The more prominent difference is that our graph-based transformer additionally uses graph convolutions to enhance local awareness. The results show the effectiveness of our multi-level structure and the graph-based transformer.

\subsection{The Relationship Between Contact Prediction and Object Reconstruction}
The contact prediction is positively correlated with the object reconstruction. To verify this assumption, we exploit the contacts estimated in the previous section for object reconstruction. 
To make sure that the gains are not due to randomness in training, we train and evaluate the model multiple times with different seeds (i.e., 123, 234, 345, and 456), and report the mean and standard deviation in the table.
Table~\ref{contact_compare} shows that the more accurate the contact prediction (larger Precision, Recall, and F1 values), the better the object reconstruction (larger F@5mm and F@10mm values and smaller chamfer distance). 
For example, the last two rows of the table show that as the F1 of contact prediction increases by 4.0\% and 4.8\% compared with the previous method, the chamfer distance decreases by 3.3\% and 5.1\% respectively for object reconstruction. 
Although the improvement percentages are not similar, the results still indicate the effectiveness of explicit contact for implicit object reconstruction.

\begin{table*}[t]
\centering
\caption{Comparison of methods using different contact features on the HO3D and OakInk datasets. The value in brackets indicates the difference between the current and the previous row.}
\label{contact_features}
\begin{tabular}{lcccc}
\toprule
Dataset & Contact Features & F@5mm$\uparrow$ & F@10mm$\uparrow$ & Chamfer Distance ($mm$)$\downarrow$\\
\midrule 
HO3D & All-one contact vector & 0.361 & 0.592 & 0.848 \\ 
HO3D & Average contact statistics & 0.369 (2.2\% $\uparrow$) & 0.604 (2.0\% $\uparrow$) & 0.779 (8.1\% $\downarrow$) \\ 
HO3D & Estimated contact features & \textbf{0.393} (6.5\% $\uparrow$) & \textbf{0.633} (4.8\% $\uparrow$) & \textbf{0.646} (17.1\% $\downarrow$) \\ 
\midrule
OakInk & All-one contact vector & 0.429 & 0.651 & 0.408 \\ 
OakInk & Average contact statistics & 0.434 (1.2\% $\uparrow$) & 0.663  (1.8\% $\uparrow$) & 0.377 (7.6\% $\downarrow$) \\ 
OakInk & Estimated contact features & \textbf{0.447} (3.0\% $\uparrow$) & \textbf{0.716} (8.0\% $\uparrow$) & \textbf{0.274} (27.3\% $\downarrow$) \\ 
\bottomrule
\end{tabular}
\end{table*}

\begin{table*}[ht]
\centering
\caption{Comparison of methods using different levels of contact states on the HO3D benchmark on F-score at 5mm and 10mm thresholds, chamfer distance (CD), penetration depth (PD), and intersection volume (IV). 
Two types of states from the contact prediction and ground truth are considered separately. The baseline IHOI is also provided for comparison.}
\label{multiho3d}
\begin{tabular}{lccccccc}
\toprule
Contact Source &  Contact Level & F@5mm$\uparrow$ & F@10mm$\uparrow$ & CD ($mm$)$\downarrow$ & PD ($cm$)$\downarrow$ & IV ($cm^{3}$)$\downarrow$ \\
\midrule
\multirow{2}*{Prediction} & Part-level & \textbf{0.393} & \textbf{0.633} & \textbf{0.646} & \textbf{0.67} & \textbf{2.91}\\
 & Vertex-level & 0.374 & 0.612 & 0.741 & 0.78 & 4.05 \\
\midrule
\multirow{2}*{Ground Truth} & Part-level & \textbf{0.490} & 0.699 & 0.655 & 0.50 & 2.15\\
& Vertex-level & 0.488 & \textbf{0.709} & \textbf{0.583} & \textbf{0.49} & \textbf{1.90}\\
\midrule
IHOI & - & 0.351 & 0.600 & 0.656 & 0.90 & 4.10 \\
\bottomrule
\end{tabular}
\end{table*}

\begin{table*}[ht]
\centering
\caption{Comparison of methods using different levels of contact states on the OakInk benchmark. Two types of states from the contact prediction and ground truth are considered separately. The baseline IHOI is also provided for comparison.}
\label{multioakink}
\begin{tabular}{lccccccc}
\toprule
Contact Source & Contact Level & F@5mm$\uparrow$ & F@10mm$\uparrow$ & CD ($mm$)$\downarrow$ & PD ($cm$)$\downarrow$ & IV ($cm^{3}$)$\downarrow$ \\
\midrule
\multirow{2}*{Prediction} & Part-level & 0.447 & \textbf{0.716} & \textbf{0.274} & 0.66 & 3.03\\
& Vertex-level & \textbf{0.450} & 0.704 & 0.288 & \textbf{0.60} & \textbf{2.70} \\
\midrule
\multirow{2}*{Ground Truth} & Part-level & \textbf{0.461} & \textbf{0.720} & 0.281 & 0.62 & 2.73\\
& Vertex-level & 0.446 & 0.711 & \textbf{0.280} & \textbf{0.55} & \textbf{2.41}\\
\midrule
IHOI & - & 0.432 & 0.658 & 0.491 & 0.75 & 4.36 \\
\bottomrule
\end{tabular}
\end{table*}

\begin{table*}[h]
\centering
\caption{Comparison on the HO3D dataset using different diffusion and scale combination methods.}
\label{ablation_diffusion}
\begin{tabular}{lcccc}
\toprule
Diffusion Method & Scale Combination Method & F@5mm$\uparrow$ & F@10mm$\uparrow$ & CD ($mm$)$\downarrow$ \\
\midrule
Nearest Neighbor Diffusion & V3 + V4 & 0.185 & 0.356 & 2.044 \\
Sparse Convolution & V3 + V4 & 0.371 &  0.614 & 0.680 \\
\midrule
Nearest Neighbor Diffusion & V1 + V2 + V3 + V4 & 0.261 & 0.475 & 1.11 \\
Sparse Convolution & V1 + V2 + V3 + V4 & \textbf{0.393} & \textbf{0.633} & \textbf{0.646} \\
\bottomrule
\end{tabular}
\end{table*}

\subsection{Ablation Study on the Contact Features}
To verify the effectiveness of the estimated contact features for object reconstruction, we remove the contact estimation module from the architecture. For a fair comparison with our method, we introduce two contact vectors of the same dimension in the baseline. One is an all-one vector, and the other is the average contact normalized to [0.1, 1] after statistics on the corresponding benchmark.
As shown in Table~\ref{contact_features}, the method with an all-one vector performs worse than other methods on both two datasets. Then, the network with the average contact achieves better results by using the contact priors of benchmarks. Finally, our method using the contact estimation module obtains more significant improvements as most metrics indicate more than 5\% improvements on both datasets, indicating that the estimated contacts could provide more flexible and efficient guidance for object reconstruction.

\subsection{Ablation Study on the Contact Level}

In this work, the predicted contact states $\boldsymbol{C_{v}} \in \mathbb{R}^{N_{v}}$ are utilized to build structured contact codes for the implicit reconstruction.
As mentioned in the main text, current vertex-level predictions are far less accurate than part-level ones, so the part-level contact states $\boldsymbol{C_{p}} \in \mathbb{R}^{N_{p}}$ are converted to vertex-level ones (i.e., $\boldsymbol{C_{v}}$) according to the fixed correspondence for subsequent experiments. 
More comparisons and analyses on the two datasets are provided as follows.

\paragraph{Ablation Results on the HO3D Dataset.}
In Table~\ref{multiho3d}, the first two rows show quantitative results for object reconstruction using part-level and vertex-level contacts estimated in the first stage, respectively.
We observe that both of them show significant improvement over IHOI~\cite{ye2022s} which does not take the hand-object contact into account, and the former outperforms the latter.
Compared with the method using vertex-level contact prediction, the part-level one could increase $\rm {F@5mm}$ and $\rm {F@10mm}$ by 5.1\% and 3.4\% and greatly reduce the chamfer distance.
In addition, the intersection volume (2.91 $cm^{3}$) of the part-level state achieves a 28.1\% improvement over the other one (4.05 $cm^{3}$).
These results can be explained by the fact that part-level predictions are more accurate.
When the contact is perfect (Rows 3-4), our method shows a more obvious advantage over IHOI across all metrics. 
For example, the part-level state largely improves $\rm {F@5mm}$ by 39.6\% and reduces the penetration depth by 44.4\% than IHOI.
What's more, though the $\rm {F@5mm}$ (0.490) is 0.002 larger than that of the vertex level (0.488), the other metrics show a clear advantage of the vertex-level model against the part-level model, which also indicates that precise vertex-level contacts can lead to better results.

\paragraph{Ablation Results on the OakInk Dataset.}
As illustrated in Table~\ref{multioakink}, all of our methods (the first four rows) outperform the state-of-the-art method~\cite{ye2022s} on all metrics. 
For models using the predicted contacts, compared with the vertex-level model, the part-level model achieves a 1.7\% improvement for $\rm {F@10mm}$ and a 4.9\% improvement for chamfer distance.
However, the penetration depth and the intersection volume of the part level are slightly worse than the vertex level, indicating that the vertex-level contacts are more efficient for interaction reconstruction.
Furthermore, Table~\ref{multioakink} shows that compared with using predicted contacts, the ground truth contacts only bring limited improvement, such as increasing the $\rm {F@5mm}$ of the part-level model from 0.447 to 0.461 and decreasing the penetration depth of the vertex-level model from 0.60 $cm$ to 0.55 $cm$.
Similar to the conclusion in Table~\ref{multiho3d}, the ground truth vertex-level contacts could lead to the best intersection volume (2.41 $cm^{3}$). 
However, its object reconstruction is not as good as the model using estimated vertex-level predictions. The main reason is that ground-truth contact annotations on OakInk are not accurate, which also brings negative impacts to the final reconstruction results.
Therefore, it is reasonable to use more accurate part-level contact predictions for the implicit object reconstruction when precise vertex-level contact annotations are not available.

\subsection{Ablation Study on the Diffusion Method}
In this work, sparse convolution is used to diffuse contact information from hand surfaces to the 3D space for hand-held object reconstruction. To verify the effectiveness of the usage of sparse convolution, we implement variants using the nearest neighbor diffusion module with simple interpolation diffusion like LoopReg~\cite{bhatnagar2020loopreg} and validate our solution in the following two aspects:

\paragraph{Efficiency.}
The method using the nearest neighbor diffusion is nearly ten times slower than that using sparse convolutions.

\paragraph{Strategy to Diffuse Contact Information.}
In our paper, the results are produced by using four contact code volumes: V1 = [32, 32, 32, 32], V2 = [16, 16, 16, 64], V3 = [8, 8, 8, 128], V4 = [4, 4, 4, 128]. We compare two strategies to diffuse contact information at different scales in Table~\ref{ablation_diffusion}. For the scale combination of V3 and V4 with the smallest resolution, the results show that the method using nearest neighbor diffusion (e.g., F@5mm = 0.185) is worse than that using sparse convolutions (e.g., F@5mm = 0.371). Similar conclusions are also valid for combinations using the four scales. In addition, it could be observed that the method using the nearest neighbor diffusion with four scales (e.g., F@5mm = 0.261) performs worse than the IHOI baseline (e.g., F@5mm = 0.351 in Table~\ref{multiho3d}) which does not use contacts. 
The results show that a naive multi-scale volume pyramid is not suitable for this task since the estimated contacts from hand surfaces are not fully learned and utilized by the network, which even brings negative effects for implicit object reconstruction.

\begin{table}[t]
\centering
\caption{Ablation study on the multi-scale contact code volumes on the HO3D dataset.}
\label{ablation_msc}
\begin{tabular}{lccc}
\toprule
Scale Combination & F@5mm$\uparrow$ & F@10mm$\uparrow$ & CD $\downarrow$ \\
\midrule
V1 & 0.327 & 0.562 & 0.77 \\
V1 + V2 & 0.348 & 0.589 & 0.701 \\
V1 + V2 + V3 & 0.386 &	0.633 &	0.684 \\
V1 + V2 + V3 + V4 & \textbf{0.393} & \textbf{0.633} & \textbf{0.646} \\
\bottomrule
\end{tabular}
\end{table}

\subsection{Ablation Study on Multi-scale Contact Code Volumes}

Table~\ref{ablation_diffusion} illustrates that the methods with the scale combination of V3 and V4 all perform worse than the ones with four scales. We further take an ablation study on the multi-scale contact code volumes as shown in Table~\ref{ablation_msc}. As the number of volumes increases, the reconstruction gradually becomes better, showing the effectiveness of full multi-scale contact code volumes.

\subsection{Ablation Study on the Contact Loss}

Our method is object model-agnostic during inference, similar to the task setting of~\cite{hasson2019learning,karunratanakul2020grasping,ye2022s}. The relevant methods use contact information in an implicit way. They introduce contact loss consisting of attractive and repulsive terms in the context of neural implicit representation, which could be compared with our proposed HOI loss $L_{hoi}$. We implement the attractive and repulsive loss terms $L_{attrep}$ based on SDF values following~\cite{hasson2019learning}. The positive and negative values of SDF represent the intersection and separation between the hand and object vertices. The core difference between $L_{attrep}$ and our proposed $L_{hoi}$ lies in that our loss could use more precise contact information by taking the estimated contact probability $c_v$ as the weight. As shown in Table~\ref{ablation_loss}, by using $L_{attrep}$, the baseline method IHOI could also be improved to 0.670 from 0.658 on F@10mm. However, the baseline still performs worse than our method (e.g., F@10mm=0.704) using the same $L_{attrep}$, showing the effectiveness of the proposed contact features $f_{c}$. 
Furthermore, our method could be improved by using the proposed $L_{hoi}$. For instance, the chamfer distance could be further improved by 19.0\% to 0.260, demonstrating that flexible and accurate contact guidance is highly beneficial to this task.

\begin{table}[t]
\centering
\caption{Ablation study on the contact loss on the OakInk benchmark.}
\label{ablation_loss}
\begin{tabular}{lccc}
\toprule
Method & F@5mm$\uparrow$ & F@10mm$\uparrow$ & CD $\downarrow$ \\
\midrule
IHOI & 0.432 & 0.658 & 0.491 \\
IHOI with $\mathcal{L}_{attrep}$ & 0.444 & 0.670 & 0.426 \\
Ours with $\mathcal{L}_{attrep}$ & 0.458 & 0.704 & 0.321 \\
Ours with $\mathcal{L}_{hoi}$ & \textbf{0.459} & \textbf{0.718} & \textbf{0.260} \\
\bottomrule
\end{tabular}
\end{table}

\section{Qualitative Results}\label{Qualitative}

\begin{figure*}[t]
\centering
\includegraphics[width=0.9\textwidth]{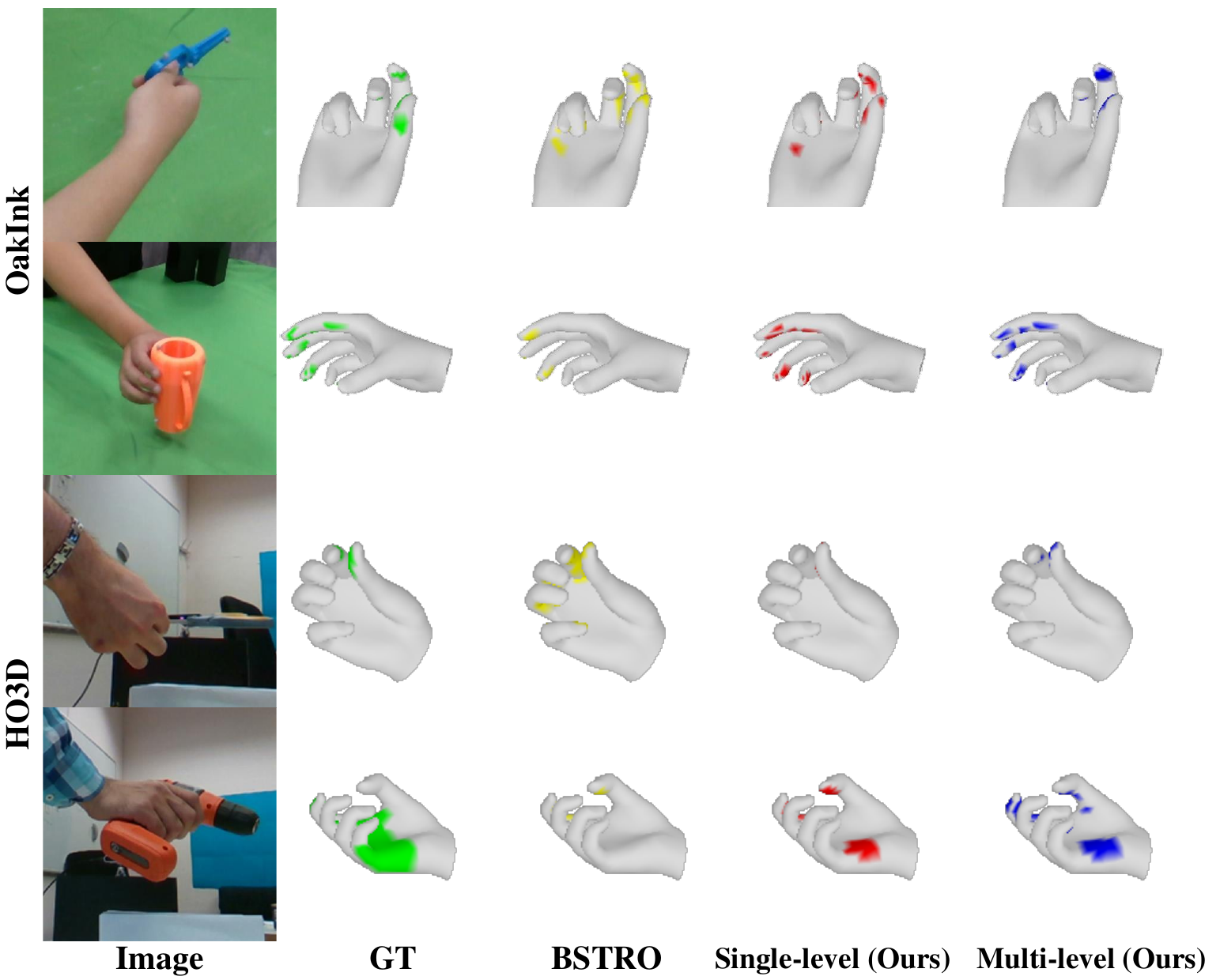}
\caption{Vertex-level contact prediction on the OakInk and HO3D datasets. For the sample whose contact regions are occluded by hands, the hand mesh is rotated 90 or 180 degrees for clear visualization.}
\label{rebuttal_contact}
\end{figure*}

\begin{figure*}[h]
\centering
\includegraphics[width=0.9\textwidth]{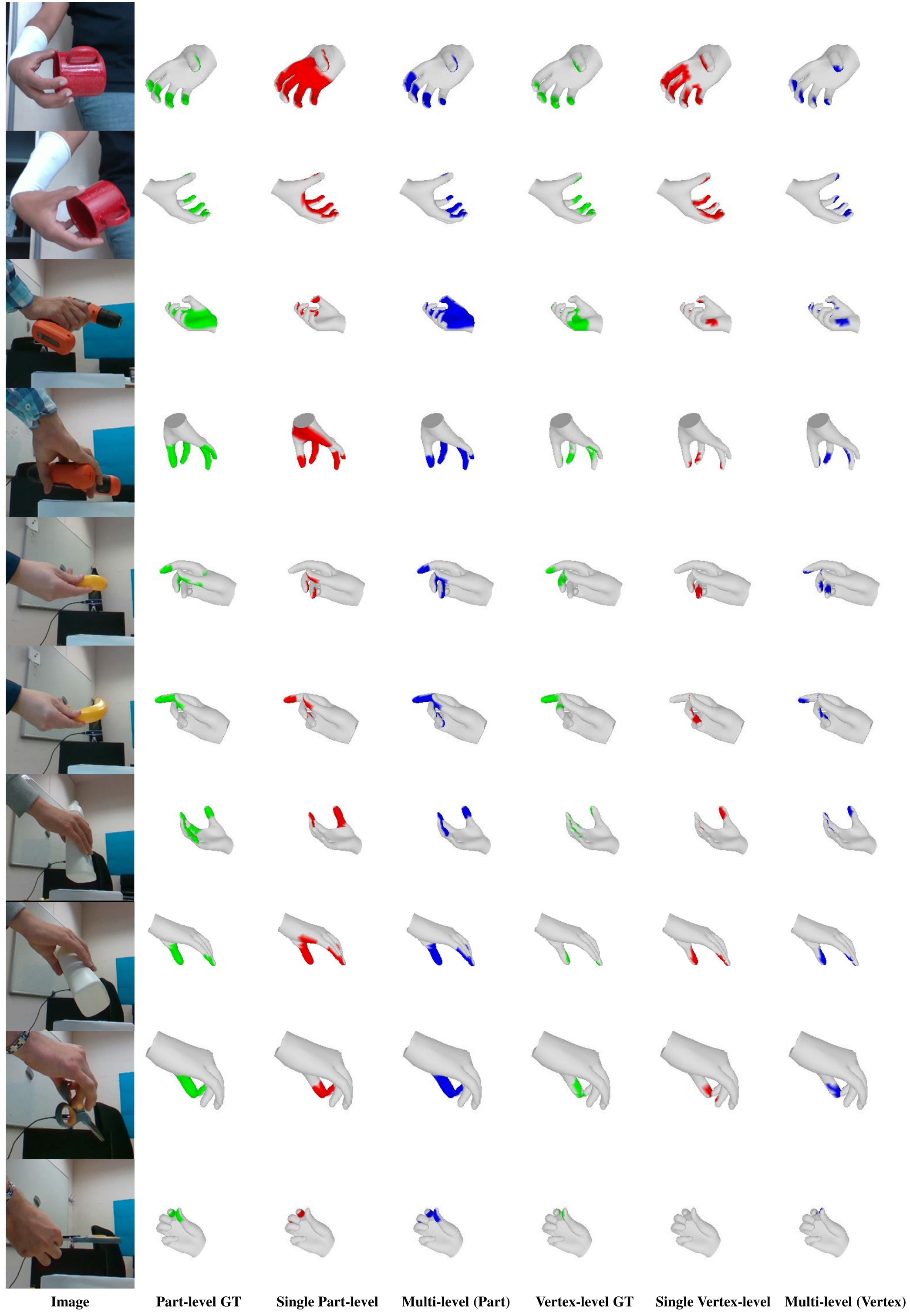}
\caption{Contact prediction on the HO3D dataset.
The ground truth area is green, the single part or vertex-level prediction is red, and the multi-level estimation corresponds to blue.
For the sample whose contact regions are occluded by hands, the hand mesh is rotated 90 or 180 degrees for clear visualization.}
\label{contactho3d}
\end{figure*}

\begin{figure*}[h]
\centering
\includegraphics[width=0.9\textwidth]{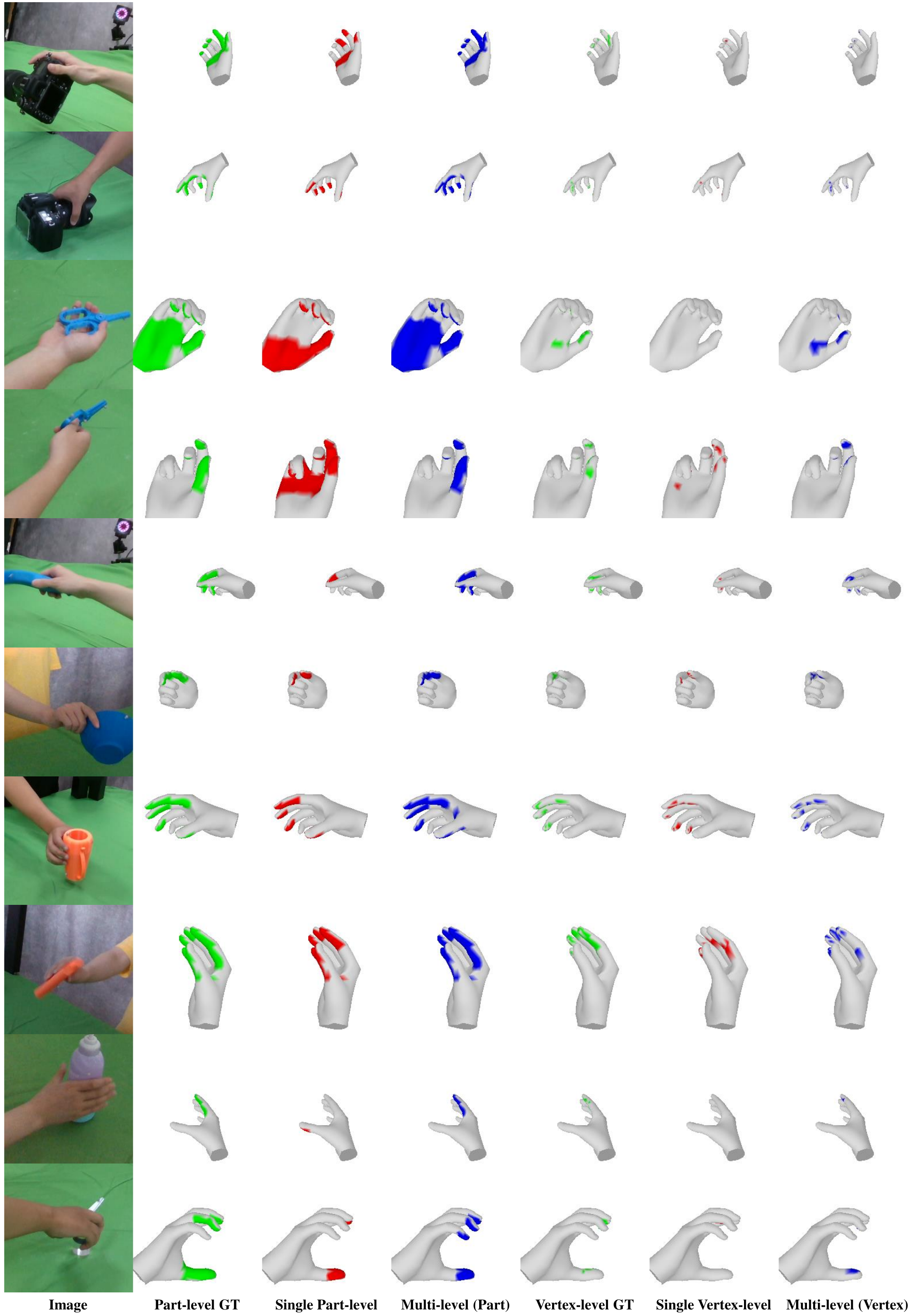}
\caption{Contact prediction on the OakInk dataset.
The ground truth area is green, the single part or vertex-level prediction is red, and the multi-level estimation corresponds to blue.
For the sample whose contact regions are occluded by hands, the hand mesh is rotated 90 or 180 degrees for clear visualization.}
\label{contactoakink}
\end{figure*}

\paragraph{Contact Prediction.}
Fig.~\ref{rebuttal_contact} shows the vertex-level contact prediction on the OakInk and HO3D datasets. 
It can be observed that the proposed methods outperform BSTRO~\cite{huang2022capturing} because the graph-based transformer additionally uses graph convolutions to enhance local awareness.
Fig.~\ref{contactho3d} and Fig.~\ref{contactoakink} show more qualitative results of contact estimation on two datasets.
The proposed method is robust to complex hand poses and occlusions from hands or objects.
Our approach can produce reasonable results even for challenging samples such as picking up a drill or using scissors.
Compared with the multi-level structure, the single-level method cannot accurately predict the contact area when dealing with occluded samples, which indicates the superiority of the multi-level architecture.

\begin{figure*}[h]
\centering
\includegraphics[width=1.0\textwidth]{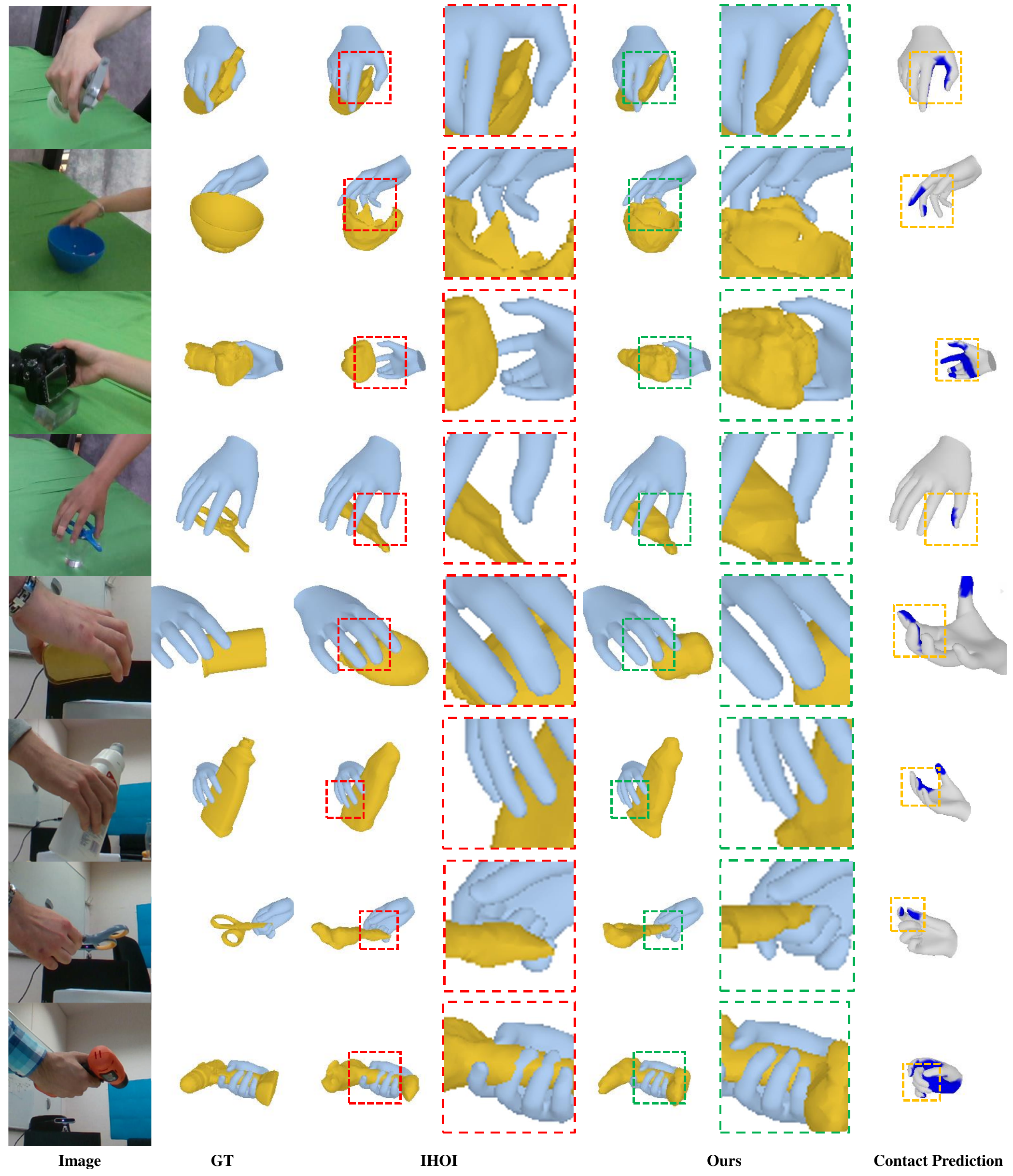}
\caption{Qualitative comparison with the state-of-the-art method on the OakInk (Rows 1-4) and HO3D (Rows 5-8) datasets. The proposed method is more robust to occlusions and unseen objects during training than IHOI.} 
\label{morecompre}
\end{figure*}

\begin{figure*}[h]
\centering
\includegraphics[width=1.0\textwidth]{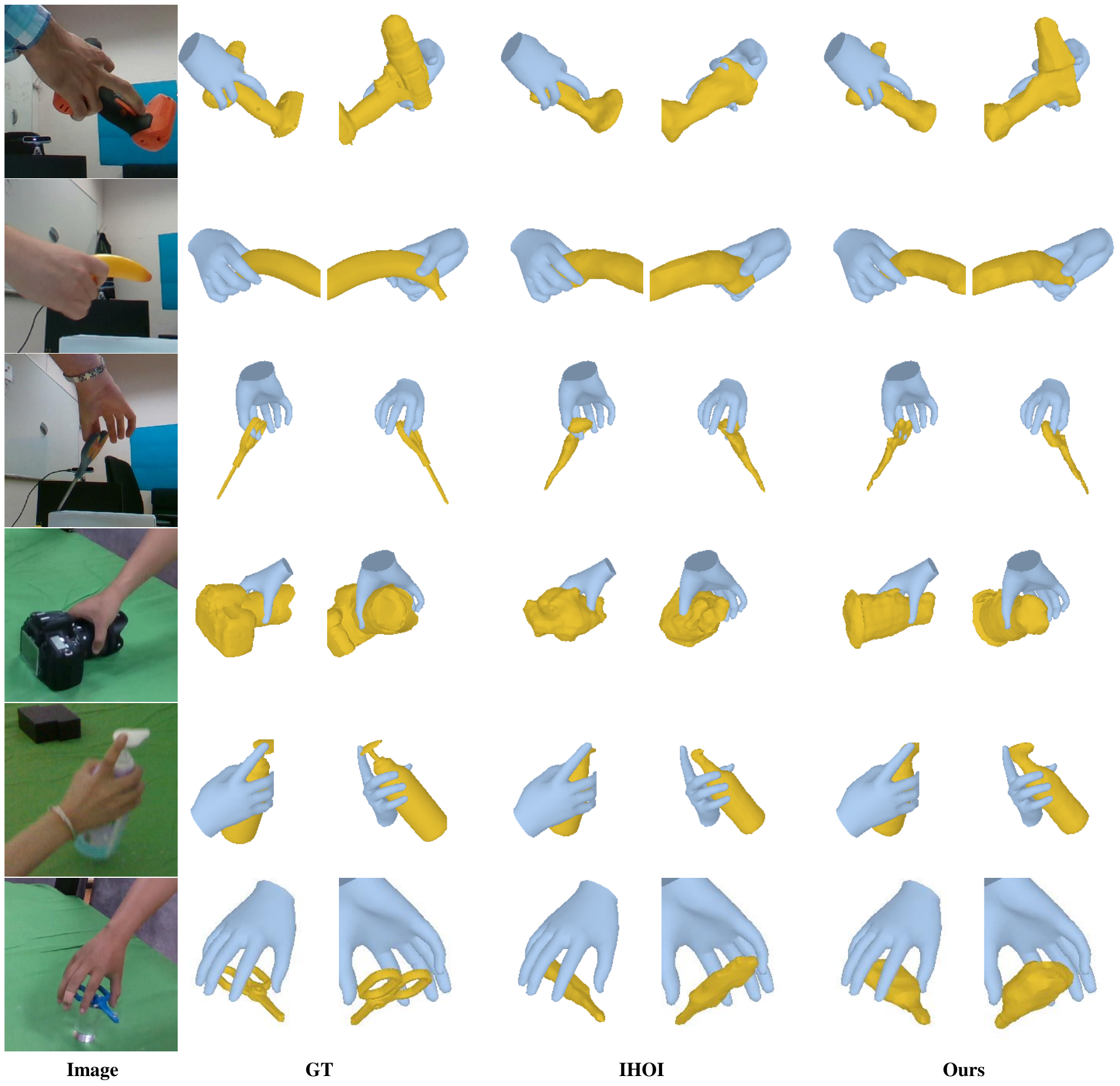}
\caption{Object reconstruction with multiple views on the HO3D (Rows 1-3) and OakInk (Rows 4-6) datasets.} 
\label{multiviewresults}
\end{figure*}

\paragraph{Object Reconstruction.}
Fig.~\ref{morecompre} shows more qualitative results with the state-of-the-art method~\cite{ye2022s}. As shown in the results of the first four rows, for unseen objects during training (\emph{i.e.}, from the OakInk dataset), the proposed method can still perform well, especially for object parts that are in contact with the hand. Furthermore, Fig.~\ref{multiviewresults} demonstrates the object reconstruction with multiple views. Since the structured contact codes provide both positional and contact information, our approach can still produce plausible results for occluded object parts.

\end{document}